\pdfoutput=1

\documentclass[11pt]{article}

\usepackage{acl}

\usepackage{times}
\usepackage{latexsym}
\usepackage{xcolor}
\usepackage{booktabs}
\usepackage{fancyhdr}
\usepackage{fullpage}
\usepackage{titlesec}
\usepackage{times}
\usepackage{url}
\usepackage{hyperref}
\usepackage{graphicx}
\usepackage{microtype}
\usepackage{wrapfig}
\usepackage{enumitem}
\usepackage{titlesec}

\usepackage[T1]{fontenc}

\usepackage[utf8]{inputenc}

\usepackage{csquotes}
\usepackage{fancyvrb}

\usepackage{amsmath,amsfonts,bm}

\def\eqref#1{equation~\ref{#1}}

\def\1{\bm{1}}

\DeclareMathAlphabet{\mathsfit}{\encodingdefault}{\sfdefault}{m}{sl}
\SetMathAlphabet{\mathsfit}{bold}{\encodingdefault}{\sfdefault}{bx}{n}

\DeclareMathOperator*{\argmax}{arg\,max}

\usepackage{algorithm}
\usepackage{algorithmic}

\usepackage{multirow}

\usepackage{xspace}
\usepackage{mdframed}
\usepackage{sidecap}
\usepackage{comment}

\usepackage{tikz}
\usetikzlibrary{calc}
\usetikzlibrary{shapes}

\newcommand{\cA}{\mathcal{A}}

\newcommand{\cH}{\mathcal{H}}

\newcommand{\cN}{\mathcal{N}}

\newcommand{\cS}{\mathcal{S}}

\newcommand{\cZ}{\mathcal{Z}}

\newcommand{\chatgames}{\texttt{chat\_games}}

\title{Steering Language Models with Game-Theoretic Solvers}

\author{Ian Gemp \and Roma Patel \and Yoram Bachrach \and Marc Lanctot \\
{\bf Vibhavari Dasagi} \and {\bf Luke Marris} \and {\bf Georgios Piliouras} \and {\bf Siqi Liu} \and {\bf Karl Tuyls} \\
Google DeepMind, London, UK \\
\texttt{imgemp@google.com}}

\begin{document}
\maketitle
\begin{abstract}
Mathematical models of interactions among rational agents have long been studied in game theory.
However these interactions are often over a small set of discrete game actions which is very different from how humans communicate in natural language.
To bridge this gap, we introduce a framework that allows equilibrium solvers to work over the space of natural language dialogue generated by large language models (LLMs). Specifically, by modelling the players, strategies and payoffs in a ``game'' of dialogue, we create a binding from natural language interactions to the conventional symbolic logic of game theory.
Given this binding, we can ask existing game-theoretic algorithms to provide us with strategic solutions (e.g., what string an LLM should generate to maximize payoff in the face of strategic partners or opponents), giving us predictors of stable, rational conversational strategies.
We focus on three domains that require different negotiation strategies: scheduling meetings, trading fruit and debate, and evaluate an LLM's generated language when guided by solvers.
We see that LLMs that follow game-theory solvers result in dialogue generations that are less exploitable than the control (no guidance from solvers), and the language generated results in higher rewards, in all negotiation domains.
We discuss future implications of this work, and how game-theoretic solvers that can leverage the expressivity of natural language can open up a new avenue of guiding language research.

\end{abstract}

\section{Introduction}
Although large language models (LLMs) have achieved remarkable performance on many natural language tasks \citep{radford_2020, jiang2023mistral, team2023gemini}, there are many studies showcasing the failures in the reasoning capabilities of such models---for example, even when the surface form of LLM-generated text appears plausible and human-like, they often fail to exhibit rational and consistent reasoning strategies \citep{jiang-etal-2020-know, turpin2024language}.
In a different area of research however, the study of rational and strategic behavior amongst agents has long been studied in the field of game theory \citep{fudenberg1991game, rosenschein1994rules}, where mathematical models and algorithms can solve for such optimal behavior allowing agents to win complex games. 
These models have applications in a wide range of fields, from social science, economics to cognitive science \citep{stone2000multiagent}, allowing us to use this formulation to build agents exhibiting intelligent reasoning strategies against competitors or cooperators in multi-agent settings.
For example, in games that require complex decision making like Diplomacy and StarCraft \citep{vinyals2019grandmaster, meta2022human}, methods that can solve for an equilibrium of a game have propelled agents to human-level performance.

However, most of these success stories are on games in the colloquial sense of the word. These are card, board, or video games that often have action spaces far smaller than the space of natural language words. Language also comes with the underlying complexities of strategizing about beliefs and intents of players over the semantics of the generated words.
In this work we ask whether there exists a generic way to combine outputs from game-theoretic models (e.g., ones that can solve for optimal equilibrium solutions) to guide language model generations in strategic games. We show how we do not have to be restricted to traditional ``games" studied in game theory, but can extend to standard dialogue tasks, where two LLMs generate natural language to communicate with each other. 
We do this by creating a mapping from natural language dialogue tasks to the formal framing of \emph{extensive-form} games, thus allowing off-the-shelf equilibrium solvers to find the optimal action an agent should be taking. When information from such a solver is then fed back into an LLM, we show how this can lead to more rational and strategic natural language generations from LLMs in dialogue games. In three different domains, we evaluate the extent to which a game-theory guided LLM outperforms a standard LLM and also analyze the effect of different types of equilibrium solvers on LLM generations.
\begin{figure*}[!ht]
    \centering
    \includegraphics[width=0.9\textwidth]{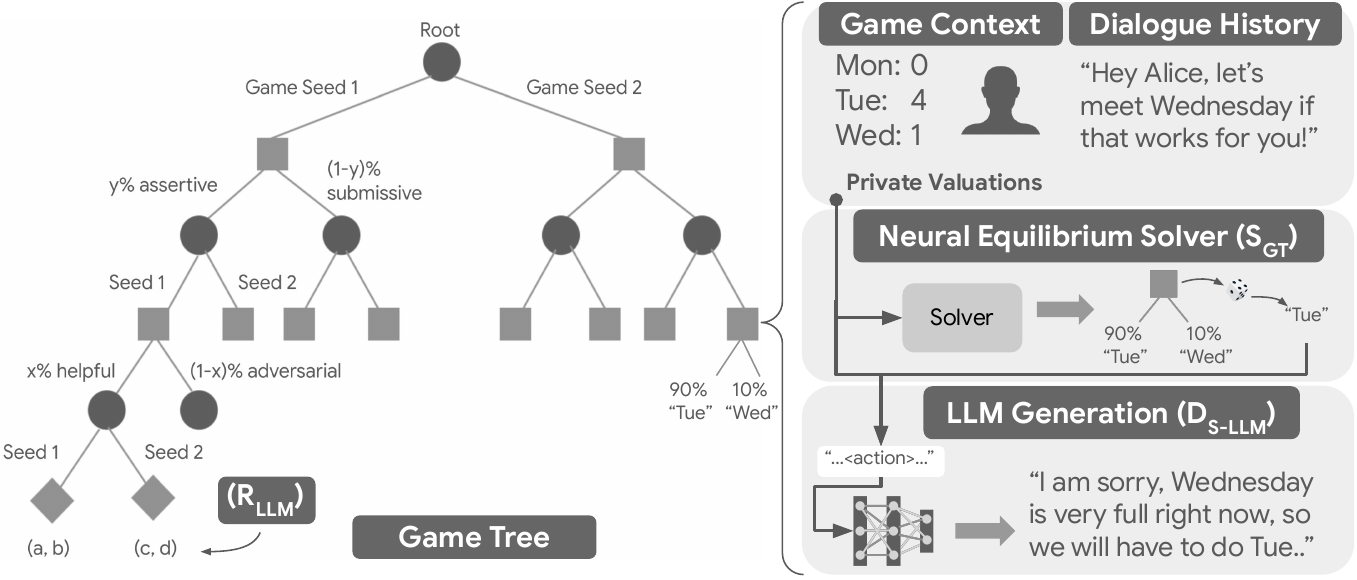}
    \caption{Figure shows an overview of our framework: on the left, an example dialogue game tree in a meeting scheduling domain, and on the right, an LLM prompted with equilibrium solver decisions. Squares denote decision points where the solver chooses actions, circles denote chance nodes where LLM responses are stochastically generated, and diamonds denote leaves (terminal states of the tree). Values below the diamonds correspond to the payoffs for player 1 and player 2, respectively.}
    \label{fig:overview}
\end{figure*}

In summary, our contributions are as follows. We first develop a binding from conversational dialogue to the language of game theory. Given this framing, we use off-the-shelf game-theory solvers to compute an equilibrium strategy. The equilibrium strategy is then passed back to an LLM as part of its context, that it can use to guide the next natural language response it generates to its opponent. We then evaluate whether the influence of the equilibrium solver does indeed lead the LLM to generate more strategic responses and achieve higher payoff in the game.

The remainder of this paper is framed as follows.
In Section~\ref{sec:framing} we introduce our framing of dialogue as a formal game by mapping to key concepts of game theoretic models. We show how game-theoretic solvers, when combined with LLMs, can work in three dialogue domains (Section~\ref{sec:methodology}) and present experiments providing empirical support for using game-theoretic models to improve LLM outputs (Section~\ref{sec:results}). Lastly, we outline related work in the area, and discuss the limitations and also implications of this work.

\section{Framing Dialogue as an Extensive-Form Game}\label{sec:framing}
Extensive-form, imperfect-information games represent one of the most general classes of games in game theory. They allow us to represent sequential (i.e., temporally-extended) interactions between players and their underlying valuations or preferences that influence the actions they take in a game. In this section we outline our formal framing of dialogue as a traditional extensive-form game i.e., a tuple $\langle \cN$, $c$, $\cA$, $\cH$, $\cZ$, $u$, $\tau$, $\cS \rangle$ and define all the necessary elements of a traditional game below.\footnote{The exact binding we present is not unique. We will explore an approach later, PSRO, that can help to modify and improve an initial selected binding automatically.}

\begin{itemize}
\item $\cN = \{ 1, 2, \cdots, n \}$ is a set of $n$ {\bf players}. There is also a special player, $c$, called {\bf chance} (or ``nature'').
\item $\cH$ is a finite set of {\bf histories}. Each history is a sequence of actions (including chance \emph{outcomes}) taken from the start of the game.
\item $\tau: \cH \rightarrow \cN \cup \{ c \}$ is a player identity function; i.e., $\tau(h)$ indicates whose turn it is.
\item $\cS$ is a set of {\bf infostates}. Each infostate $s$ represents a partition of $\cH$ such that each $h \in s$ cannot be distinguished by players other than $\tau(s)=\tau(h)$ for all $h \in s$. 
In our games, we implicitly define these partitions by defining what information is public versus private to player $\tau(s)$, and a player's infostate therefore includes observable dialogue history, action history and private information. In each of our games, all sent messages are appended to a public thread making the entire dialogue history public knowledge. Note this still omits certain private information like each player's action, e.g., the precise \emph{tone} a message is written in although this might be approximately inferred by a player.  We only require each player have knowledge of their own action history to maintain {\bf perfect recall}\footnote{Without this property, solving a game can be substantially more expensive computationally.}.
\item $\cA$ is a finite set of \emph{infostate-dependent} \textbf{actions} that can be taken. In the case of dialogue, an LLM can receive a game context and natural language instructions as input; an instruction in this case constitutes the action (strategy) to be taken and is assumed to come from a finite instruction set, e.g., \emph{use a tone that is} \{``assertive'', ``submissive''\}. Note, one could extend the action set to include a choice of which LLM (e.g., out of a set of models of different sizes) to call as well; the only requirement is that this set remain finite. 
\item $\cZ \subseteq \cH$ is a set of {\bf terminal histories} determined by either a limit on the allowable tree-depth (we only allow a finite number of messages per player) or by an LLM tasked with determining whether a conversation has ended e.g., a deal has been struck and no further conversation is required.
\item $u: \cZ \rightarrow \Delta_u^n \subseteq \Re^n$ with $\Delta_u = [ u_{\min}, u_{\max} ]$ is a utility (or payoff) function assigning each player a payoff at the end of the game. We construct a prompt specific to the domain at hand and let an LLM quantify the reward to each player following previous work~\citep{kwon2023reward,wei2022chain}.
\end{itemize}

\paragraph{Solution to Games}

Given this definition of a game, we can attempt to solve for an optimal strategy for each player. Note that in games, the behaviour of players influences the payoffs of the other players. The interactions between agents may be purely competitive, purely cooperative, or a mixture between the two. In  any case, the each player's optimal strategy should be in \emph{equilibrium} with the others: no player should have an incentive to deviate away from their strategy.

As an example, the most famous solution to a two-player zero-sum game is John von Neumann's minimax solution \citep{neumann1928_normal_form}. At this equilibrium solution, the resulting actions are unexploitable, and if there are multiple solutions, they are interchangeable. The Nash equilibrium \citep{nash1951_neq} is a famous generalisation of this solution concept to many-player games---it provides a strategy profile (meaning a mixed strategy for every player in the game) under which no player has an incentive to unilaterally deviate. Many other rich solution concepts have been defined \citep{aumann1974_ce,hannan1957_cce} and we use \textbf{equilibrium solvers} to solve the games we define.

\paragraph{Our framework} Given this formulation above, we can frame a dialogue task between two agents as a two-player extensive-form game. The players are LLMs that can generate natural language responses. The strategies that an equilibrium solver guides an LLM towards can be distributions over instruction sets that the model should condition on (along with other useful dialogue history or game context) when producing a response. Separate LLMs can act as reward models to assign payoffs to the agents at the end of the game by evaluating the responses generated. We outline details of this methodology in the next section.

\begin{figure*}[ht!]
    \centering
    \includegraphics[width=0.98\textwidth]{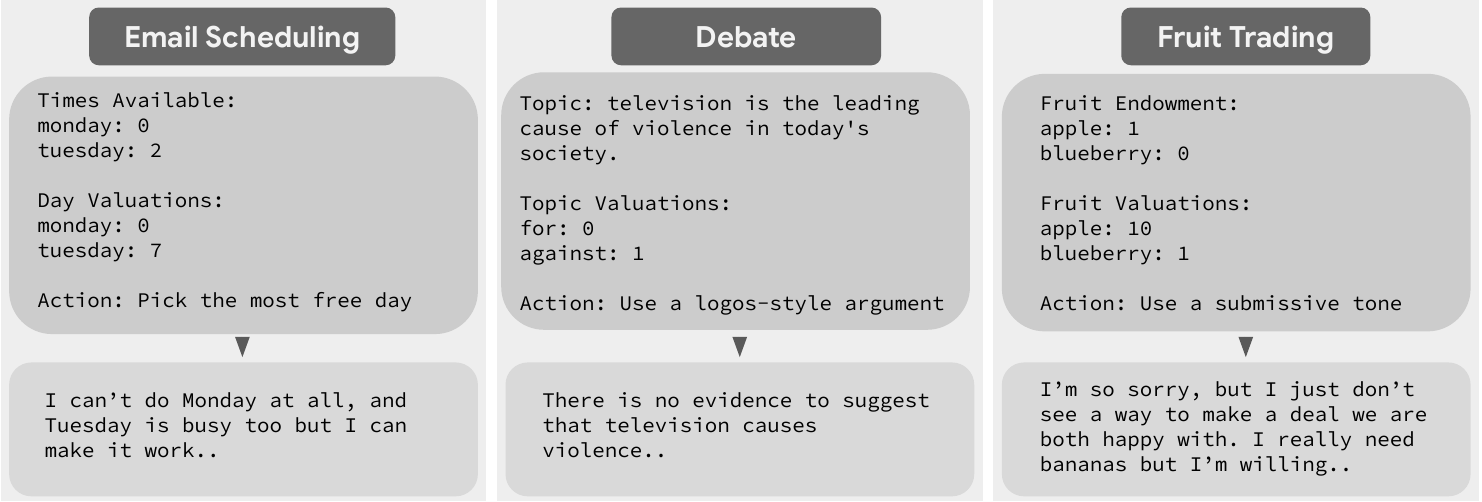}
    \caption{Illustration of the three dialogue domains we consider: an email scheduling task on the left, a debate task in the centre, and a fruit trading task on the right.}
    \label{fig:domains}
\end{figure*}

\section{Experimental Methodology}\label{sec:methodology}

\subsection{\bf Models} 
We use two autoregressive large language models (LLM) from the PaLM model family \citep{anil2023palm}. The models are pretrained on on the PaLM-2 pretraining corpus and then used at inference-time to generate dialogue utterances. Specifically, for a given domain, a dialogue LLM (henceforth $D_{LLM}$), conditions on a prompt that contains information about the domain, and a few turns of dialogue that it is then required to respond to. We use a solver, $S_{GT}$, described in Section \ref{sec:algorithms} to compute the optimal action to be taken in a given dialogue infostate and use this to guide the $D_{LLM}$ generations. Our baseline LLM takes the special token ``Any'', representing an uninformative input; we use $D_{LLM}$ to refer to this baseline model from now on. When using guidance from a game-theoretic solver, the prompt includes the solver's suggested instruction string that therefore influences its response. We provide example prompts in Appendix \ref{app:domains:fruit_trading_prompt}.
We also use LLMs as reward models $R_{LLM}$, where the LLMs are given few-shot, chain-of-thought-prompting samples to calculate payoffs as shown in Appendix \ref{app:reward}.

\subsection{\bf Dialogue Game Domains}

We consider three different dialogue domains shown in Figure \ref{fig:domains} that require strategic reasoning between two agents. These games are procedurally generated by defining elements in the game (e.g., items to trade, names of players) or characteristics to exhibit (e.g., different tones) that can form the initial game context or instruction set for the dialogue. The LLM gets prior information about the game domain along with the dialogue history as part of its prompt and is required to generate a response. When it is given guidance and additional input from a game theory solver (e.g., \emph{use a tone that is} \{tone\}), we aim to evaluate if this solver input modulates its response such that the augmented generation is significantly different or more strategic than the baseline LLM response.

\paragraph{Scheduling Meetings}
In this domain, players are required to schedule a meeting in a multi-turn negotiation setting.
Each player begins with a set of allowable days of the week they are available to meet in (e.g., ``monday'' and ``tuesday'') and have non-negative valuations over each day of the week. Both of these pieces of information are private to the players. Players can choose to reveal this information in their natural language generations and their actions here are the days of the week on which they propose to meet. The game rewards players according to how much they value the agreed upon day, and both players receive zero reward if no agreement is made. Here, the equilibrium solvers can guide LLMs on which action (e.g., day of the week) might be the most optimal given the context history of the game and the players' valuations.

\paragraph{Trading Fruit}

We use a fruit trading domain similar to~\citep{chawla2023selfish}, where each player begins with a private endowment of types of fruit as well as private valuations over the items. Players are rewarded by the difference in value between their basket after the trade and that before the trade. Previous work on this game theoretic task has argued that personality or tone has an impact on negotiations in natural language and we use the equilibrium solver guidance to modulate the tone of the LLM generations (e.g., ``submissive'', ``assertive''). There are four possible tones to choose from and we outline the actions and example prompts from this game in Appendix \ref{app:reward}.

\paragraph{Public Debate}

We also consider the domain of debate \citep{brown2023scalable, irving2018ai}, where LLMs are presented with an argument topic and directed to argue for the statement or against. We consider a list of standard debate topics (Appendix \ref{app:domains:debate-topics}). The action space here is the argument style: namely logos, ethos or pathos, taken from Aristotle's three modes of persuasion that is widely used and studied in debate forums \citep{aristotle1991translated}. Agents in this game have to argue either for or against the given topic, and the game rewards players by scoring the debate with either a $0$ or a $1$ depending on the validity of the argument as judged by an LLM reward model.

\subsection{\bf Algorithms to Solve Games}\label{sec:algorithms}

Framing dialogue as a game allows us to ``solve'' dialogue using a variety of equilibrium solvers.

\paragraph{CFR} For example, Counterfactual Regret Minimization (CFR) is an algorithm for solving imperfect information games~\citep{zinkevich2007regret}. In CFR, each player measures how much they could have gained by switching to a different strategy at a given infostate and then attempts to minimize this regret. An important result of players using regret-minimizing algorithms is that their time average play converges to a coarse-correlated equilibrium (CCE)~\citep{gordon2008no}.

\paragraph{PSRO}
Another algorithm is the Policy-Space Response-Oracle (PSRO) method which alternates between two steps.
In one step \textbf{Policy-Space}, a set of candidate player policies are evaluated in head-to-head matchups against each other and the outcomes (e.g., success rates of each policy) are recorded in a \emph{normal-form} game payoff matrix. The equilibrium of this matrix game is then calculated. In the other step \textbf{Response-Oracle}, each player computes a best response to this equilibrium strategy resulting in a new candidate policy, which is then added to the candidate set used in the other step. In the dialogue game, the actions are instruction prompts. An \emph{approximate best response} can be generated by sampling new random instruction prompts, evaluating them against the current equilibrium, and then returning the one with highest expected payoff. We provide pseudocode for an approximate best response operator in Algorithm~\ref{alg:abr:shotgun}. PSRO offers a strength beyond CFR in that it is able to generate new instructions (e.g., ``sycophantic'' could be generated in the Response-Oracle step) beyond the original instruction set. In that sense, PSRO is able to approximate equilibrium strategies in the open-ended universe of dialogue. Pragmatically, it is bound by the finite number of Response-Oracle steps it takes and can never truly explore the infinite landscape of language.

\section{Evaluation}\label{sec:results}
Given the framework described in the above two sections, we now empirically evaluate the influence of equilibrium solvers on LLMs. We empirically evaluate the components $D_{S-LLM}, R_{LLM}, S_{GT}$ separately: the ability of an instruction to faithfully influence an LLM's generated response, the ability of an LLM to compute a player's payoff given the dialogue history and game context, as well as the game-theoretic solver's impact on steering an LLM to generate better dialogue. We also qualitatively evaluate the dialogue generations to assess the difference in outputs in Appendix~\ref{app:steering:example_generations}.

\paragraph{Do LLMs follow game-theoretic solver outputs?}

We first evaluate whether the generations from an LLM \emph{could} reflect the guidance from the solver. For example, when the suggested equilibrium action is to sound ``submissive", can we evaluate the responses from the LLM to empirically assess how often they do actually follow this instruction to modulate the tone of their generations.

To evaluate this, given an LLM-generated message $m$ conditioned on a prompt including an instruction to take an action $a \in \mathcal{A}$, we ask whether $a$ is actually the most likely action in $\mathcal{A}$ conditioned on $m$ using a held-out model $P$, i.e., $a \in \argmax_{z \in \mathcal{A}} P(z \vert m)$. We provide details of prompt instructions for the classifier $P$ in Appendix~\ref{app:steering:accuracy}. For example, in the debate domain, when the LLM is conditioned on instructions guiding it towards ``logos'', ``ethos'', or ``pathos'' arguments, the model $P$ recognizes the given instruction as \emph{most likely} from the generated response on $75/100$ instances. Full results are reported in Appendix~\ref{app:steering:accuracy}, Table~\ref{tab:steering:debate:s} along with results on other domains and model sizes.

\paragraph{How good is an LLM as a reward model?}

We consider three different dialogue outcome types to assess the fidelity of $R_{LLM}$ in the fruit trading and meeting scheduling domains. \emph{Valid} considers valid agreements only where each player agrees to an outcome that is possible given their private information (fruit in their basket or days they are out-of-office); \emph{Rejected} considers outcomes where at least one of the players has rejected reaching an agreement all together; \emph{Incomplete} considers dialogues that ended before an agreement had been reached (e.g., due to a limit on the tree depth).

For each outcome type, we specify a two-turn dialogue template which we complete with known quantities that allow us to compute the ground truth values of trade outcomes. For example in fruit trading, \emph{Valid} trades use the following template:

{\small
\begin{verbatim}
Hi {receiver}, I would like to trade you
{num_give} {fruit_give} for {num_receive}
{fruit_receive}. Would you like to trade
with me? Best, {sender}
...
Hi {sender}, Yes, I would like to make
that trade with you! Best, {receiver}
\end{verbatim}
}

\emph{Rejected} and \emph{Incomplete} use the same sender message, but the receiver replaces the content and closing with the following two messages respectively:

{\small
\begin{verbatim}
No, I do not want to do this trade with
you. Thanks though,
\end{verbatim}
}
and
{\small
\begin{verbatim}
No, but would you accept a different
trade? Best,
\end{verbatim}
}

We consider $1000$ different randomly generated scenarios for each outcome type; in particular, the names of the players, their private information (e.g., fruit inventory and values), and proposed agreements are generated randomly such that they satisfy the outcome constraint. There are two players, therefore, each outcome generates $2000$ samples for calculating reward errors.

We report reward error according to two different metrics in Appendix~\ref{app:reward:quant_eval}, Table~\ref{tab:reward_eval}. \emph{Norm} reports the average absolute reward error divided by the range of possible rewards. \emph{Sgn} reports the frequency that the reward model returned a reward whose sign ($+/-$) did not match ground truth.

\begin{table}[!ht]
\centering

    \begin{tabular}{||c||c|c||}	
     \hline
      \bf Domain & \bf NC $\downarrow$ & \bf CFR Gain $\uparrow$  \\
     \hline\hline

      \bf Debate & 0.024 & 0.106 \\  %
      \bf Meeting Scheduling & 0.010 & 0.037 \\  %
      \bf Trading Fruit & 0.009 & 0.038 \\  %
    \hline
    \end{tabular}
	\caption{Average NashConv (NC) and CFR Gain for debate, meeting scheduling, and trading fruit domains, computed over $200$ procedurally generated games. 
	NashConv is a pseudo-distance to the Nash equilibrium measuring how much players gain by deviating from $D_{CFR-LLM}$. CFR gain measures gain from switching to $D_{CFR-LLM}$ from $D_{LLM}$.
	}
	\label{tab:cfr_eval}
\end{table}

We find that using the larger PaLM2-L model significantly reduces $R_{LLM}$ error relative to the smaller PaLM2-S and to an absolute level that is often less than single digit percentage error in each domain. Therefore, we use PaLM2-L for the remainder of our experiments.

For a more qualitative assessment, we provide both positive and negative example demonstrations in Appendix~\ref{app:reward}.

\paragraph{Are LLMs more strategic under the influence of solvers?}
Now that we have confirmed that LLMs do follow instructions a significant portion of the time and that the payoffs computed by the LLMs can be sufficiently accurate, we aim to quantify how often incorporating the suggestions of a game-theoretic solver helps LLM generations. We use OpenSpiel's~\citep{LanctotEtAl2019OpenSpiel} CFR solver to obtain the equilibrium strategies. Specifically, we wish to evaluate the improvement of $D_{CFR-LLM}$ over the baseline $D_{LLM}$ that receives the uninformative instruction to use ``Any'' action. In other words, the baseline LLM always selects the ``Any'' branch at every decision node in the tree whereas $D_{CFR-LLM}$ selects the branch based on the equilibrium strategy computed.

We consider two scenarios. In the first, both players are baseline LLMs. We then ask how much would either player gain if they switched to $D_{CFR-LLM}$? We call this \emph{CFR Gain}. In the other scenario, both players select $D_{CFR-LLM}$. How much would either player gain if they switched to any other strategy (including $D_{LLM}$)? We call this \emph{NashConv} (NC). If CFR Gain is high relative to NashConv, we say that $D_{CFR-LLM}$ (approximately) satisfies the conditions for an \emph{evolutionarily stable strategy} (ESS)~\citep{smith1979game}.

\begin{figure*}[ht!]
    \centering
    \includegraphics[width=0.89\textwidth]{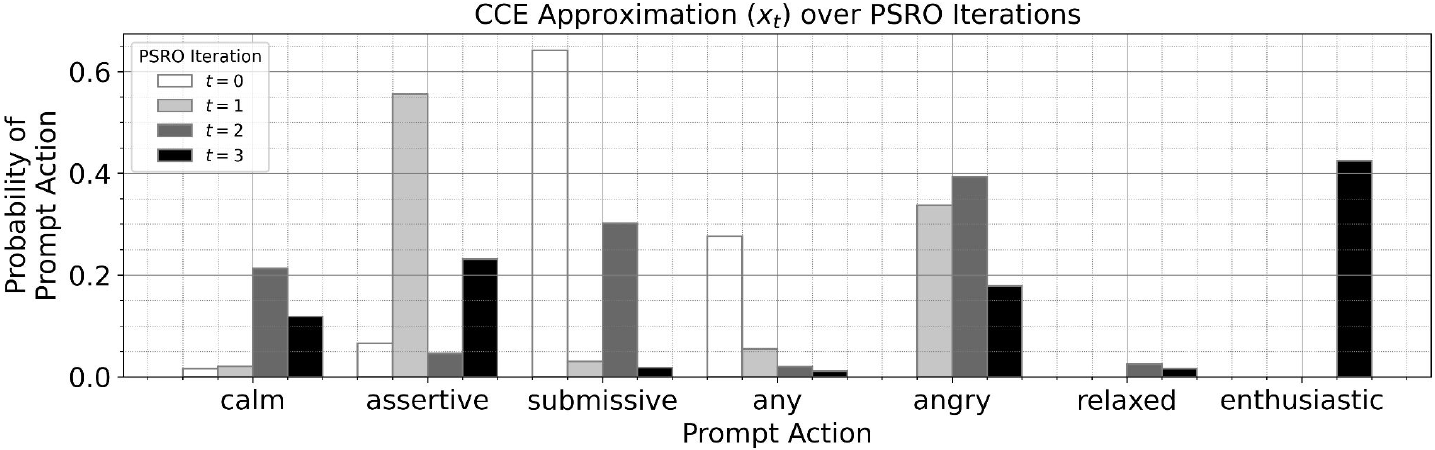}
    \caption{Progression of PSRO on the fruit trading domain. PSRO begins with the first four candidate prompts (``calm''\textemdash``any''). The equilibrium over these prompts is diplayed along with each subsequent equilbrium over the growing candidate set. Recall, each new candidate action was an approximate best response at $t=1$ to the previous candidate set (e.g., ``angry'' was a best response to the equilibrium over ``calm''\textemdash``any'') at $t=0$ shown in white.}
    \label{fig:psro:fruit_trading}
\end{figure*}

For both metrics, Table~\ref{tab:cfr_eval} shows that the LLMs benefit from following the CFR solver's guidance.
According to the criteria, $D_{CFR-LLM}$ is an approximate ESS.

\paragraph{Open-ended Strategic Dialogue}
We then consider the PSRO algorithm that alternates between solving for an equilibrium of the game and then approximating a best response to this equilibrium. In Figure~\ref{fig:psro:fruit_trading} ($t=0$), the fixed-action equilibrium distribution over the initial action set \{ ``calm'', ``assertive'', ``submissive'', ``any'' \} is reported for the fruit trading domain. We solved for this equilibrium using replicator dynamics~\citep{weibull1997evolutionary}, an algorithm from evolutionary game-theory with strong connections to regret-minimizing algorithms.
The remaining PSRO iterations ($t > 0$) generate approximate best responses (``angry'' then ``relaxed'' then ``enthusiastic'') along with equilibrium approximations over each new support.

The action ``submissive'' is initially the most probable at equilibrium\textemdash potentially a tactic to exploit another LLM by eliciting a sense of responsibility or fairness~\citep{park2023ai}.
The time steps $t > 0$ in Figure~\ref{fig:psro:fruit_trading} demonstrate the algorithm's ability to grow and refine the game's original action space by introducing novel instructions into the LLM's repertoire. We believe this feature of PSRO could be helpful in extending beyond, e.g., a negotiation expert's initial list of suggested tactics to discover new strategic instructions.

PSRO can also be generalized to focus on alternative solution concepts. In the example above, we considered replicator dynamics which converges to a CCE. However, Nash also proposed a specific solution to bargaining problems; we repeat the experiment using Nash bargaining in Appendix~\ref{app:psro}.

\paragraph{Can the optimal policies constructed by game-theoretic solvers generalize to new domains?}

Lastly, we evaluate whether the solver's equilibrium strategies can serve as optimal target policies that can be learned by a model. If possible, this imitation model, $D_{IM-LLM}$, could generalize to new domains. $D_{IM-LLM}$ can now also serve as the baseline for a new round of improvement.

We consider $200$ procedurally generated games, and use $10$ iterations of the CFR solver to approximate an equilibrium for each game. We then save T5~\citep{raffel2020exploring} vector embeddings of each information state along with the optimal equilibrium policy returned by CFR for that infostate. This constitutes a length-$768$ string embedding paired with a distribution over $\vert \mathcal{A} \vert$ actions (instruction sets) for each infostate in each game. We call the collection of such pairs our \emph{imitation dataset}.
With this dataset, it is then straightforward to train a neural network policy (also called neural equilibrium solver) to predict the equilibrium probabilities conditioned on the infostate observation. We find that a two-layer MLP trained over this dataset is sufficient to learn this distribution (see Appendix~\ref{app:imitation} for full architecture and training details). At test time, at any given turn in a dialogue, the dialogue history and game context are encoded as a (T5) vector observation; this observation is subsequently fed into our trained (MLP) neural equilibrium solver which predicts a distribution over instruction sets; we then sample a single instruction set from this distribution and pass it as input to an LLM along with the dialogue history and game context. This $D_{IM-LLM}$, outlined in Figure~\ref{fig:overview}, is then compared on held out games against the baseline $D_{LLM}$.

For evaluation, we consider a \emph{meta}-game in which two players must choose to elect a single LLM to carry out the dialogue on their behalf, e.g., negotiate a fruit trade. If the options available to these players are $D_{IM-LLM}$ and $D_{LLM}$, the approximate (coarse-correlated) equilibrium strategy is to select $D_{IM-LLM}$ $\approx 76\%$ of the time. Interestingly, in the meeting scheduling domain, this relationship is reversed, and $D_{IM-LLM}$ is chosen $\approx 44\%$ of time. We suspect that the T5 embedding of the dialogue history contains relatively little useful information for predicting the best day of the week. In contrast, it is plausible that the T5 embedding contains relatively more information about the \emph{tone} of the dialogue thus far which should be informative in selecting the subsequent ``tone'' to deploy in the fruit negotiation.

\section{Related Work}\label{sec:related_work}
There are two bodies of work most relevant to our focus. The first is on enhancing the in-context reasoning abilities of LLMs in strategic interactive settings. The second is on enabling LLMs to reason strategically in multi-agent settings by employing game-theoretic tools. We situate our work between these two areas and outline the relevant literature from both below.

\paragraph{In-context reasoning capabilities of LLMs} Existing LLMs have been shown to exhibit better reasoning capabilities when this is explicitly taught to them, either at inference time through in-context learning, ``scratchpad'' approaches \citep{wei2022chain, nye2021show}, at training-time with data containing reasoning traces of humans \citep{rajani2019explain, shwartz2020unsupervised}, or even on their own reasoning traces \citep{zelikman2022star}. Additionally, \citet{gandhi2023strategic} develop an automated ``prompt compiler'' that constructs demonstrations for LLMs to solve games and \cite{patel2021mapping} teach models to learn concepts in grounded domains. However, it is worth noting that even methods like chain-of-thought are often unreliable and unfaithful to the generations of models and susceptible to small variations in inputs that drastically alter behavior \cite{turpin2024language}.

\paragraph{Game-theoretic measures to improve LLMs}
There is a substantial line of work focused on allowing LLMs to strategically interact with one another, either in natural language or over a space of actions in a game \citep{fried-etal-2023-pragmatics}. In games that require agents to communicate with one another (e.g., Diplomacy \citep{meta2022human}) or even standard games that do not require explicit communication but could benefit from natural language hints (e.g., Minecraft-like games \citep{fan2022minedojo, rodriguez2022rlang}) there have been improvements in performance by using language-guided agents for better reasoning. \citet{lewis2017deal} consider end-to-end training of agents in negotation dialogues, but with several differences to our setting and approach. They assume a single game setting with a hard-coded reward model, item valuations are observable by all agents, and training is performed with a single best-response step due to challenges of convergence (a known pathology within computational game theory). In contrast, our approach is general to many dialogue settings, allows flexible, textual descriptions of a reward models, models private information of co-players (\emph{imperfect-information}), and enables the use of principled, convergent game-theoretic solvers; the extensions we propose align well with the desiderata posed for future work by~\citet{he2018decoupling}\textemdash avoid ``laborious dialogue state design and annotation'' and ``connect with game theory''.

\section{Conclusion}\label{sec:conclusion}
Our work investigates the extent to which natural language generations from LLMs can be steered towards more strategic behavior by game-theoretic solvers. This is motivated by the fact that current LLMs do not exhibit coherent and intelligent reasoning strategies, and also that game theoretic algorithms that could solve such problems have so far only had a limited impact on language agents. 
To address this, we formulate a framework to map natural language dialogue games to a formalism that equilibrium solvers can find solutions over, and show that across three dialogue domains, this improves the LLM generations compared to a baseline LLM that does not have access to game-theoretic solvers.
By doing so in this work, we open the door for a broad body of game theory and multi-agent research---not just on algorithms but also solution concepts and principled strategic reasoning---to pour into the AI guided interactions that pervade humans' daily life. Especially since these conversational models are already becoming a part of users' lives it is imperative that we begin to model and study these interactions to enable large language models to exhibit clear and interpretable reasoning strategies.

\section*{Limitations}
Our framework suggests that it is possible to use game theoretic models to guide the reasoning of language models in dialogue games. Given the large body of work that finds that LLMs exhibit inconsistent reasoning strategies, we see that a good solution to this problem is to offload the strategic reasoning component to a game-theoretic solver that can guide the LLM to generate language that allows it to win against an opponent. 
Moreover, we show how this framework does not have to be limited to classical games historically studied in the field of game theory. When standard natural language dialogue tasks are framed as a game, we can use existing game-theoretic solvers over such games. The solver outputs can then be used to instruct an LLM how to behave in a way that is strategically optimal. It is worth noting however, that our current approach only guides LLM generations at inference time, rather than incorporating characteristics of game-theoretic reasoning into the training or fine-tuning of LLMs. Future work that incorporates these strategies into the learning process of LLMs (e.g., as part of the objective function of a language model) can further allow more complex reasoning of these models.

However, we also wish to outline the several limitation of this framework.
Transitions in the dialogue game we pose are extremely expensive. This is because LLM inference is expensive, both computationally and financially and given the large game trees that need to be created, these calculations can very quickly become an intensive resource-consuming operation. If we want to approximate equilibria or search these game trees efficiently, we will need to push scalable game-theoretic algorithms, e.g.,~\citep{burch2012efficient}, to new heights. Additionally, our game-theoretic models are currently limited in several respects: players are assumed to have the same payoffs and action spaces and several assumptions are made on the exact structure of interactions, that is different from many tasks in the real world.
Lastly, given the illogical and incorrect reasoning patterns employed by LLMs that are not grounded in the real world~\citep{agnew2024illusion, fried-etal-2023-pragmatics} it is up for debate whether their fidelity is sufficient to draw conclusions on real world interactions. Using these models as black-box operators or autoraters for evaluation is often not 100\% reliable.
If these limitations are addressed, this implies a clear path forward towards guiding language models with "optimal" strategies found by game-theoretic models to pave the way for more intelligent language model agents.

\section*{Ethics Statement}

Strategic dialogue agents may be more ``rational'', but even in well-intended settings, it is known that seemingly benign behavior can lead to poor outcomes for the group, i.e., high \emph{price of anarchy or stability}~\citep{nisan2007agt}. Moreover, recent work has found that some of the beneficial social norms humans have developed can collapse when people are allowed to leave their coordination decisions to AI assistants~\citep{shirado2023emergence}. In more nefarious cases, strategic agents can exploit others. We should aim to create agents that are both rational, but also generate high welfare. By exploring these interactions in simulation, we can learn how to constrain and regulate agents toward more human-aligned behaviors. Designing algorithms to discover optimal (e.g, max-welfare) equilibria in $n$-player, general-sum games is an active area of research that can help guide the development of more ethical agents.

\paragraph{Societal Impact}
If the result of this work is that LLMs are, for example, ``assertive'' $x\%$ of the time and ``submissive'' $y\%$ of the time, how can we measure and/or predict their effect on human dialogue or sentiment in society at large? When the car was invented, it took decades to mass produce and gain adoption by a significant percentage of the population. LLMs, being a digital technology, could see widespread adoption only a few years after their invention. In contrast to the adoption of cars in the early 20th century, we have the capability to digitally simulate and forecast the impact of large language models. For example, if LLMs are more assertive on average than the human population, will they draw human society towards exhibiting more assertive personalities~\citep{Baumann_2020}? If LLMs are more rational and forward thinking, will these strategies similarly influence humans interacting with these systems? It is important to forecast the equilibria of these large techno-societal changes before they happen.

\bibliography{anthology,custom}
\bibliographystyle{acl_natbib}

\newpage
\onecolumn
\appendix

We provide additional details on domains in Appendix~\ref{app:domains}, quantitative evaluation of LLM instruction following with example responses in Appendix~\ref{app:steering}, quantitative and qualitative evaluation of LLM reward models in Appendix~\ref{app:reward}, additional best response algorithms and PSRO results in Appendix~\ref{app:psro}, additional imitation results in Appendix~\ref{app:imitation}, and an example code block using our codebase in Appendix~\ref{app:chatgames}.

\section{Domains}\label{app:domains}

Here we provide additional details on the debate domain and an example prompt that the LLM might see when being steered by the action suggested by a game-theoretic solver in the fruit trading domain.

\subsection{Debate Topics}\label{app:domains:debate-topics}
We source a list of debate topics that are potential sources of contention and therefore areas to practise agreeing and disagreeing on such topics. We provide examples from the list of topics below.

\begin{itemize}\itemsep0em
    \item Breakfast is the most important meal of the day.
    \item Swimming in the ocean is better than swimming in a public pool.
    \item Alcohol should be illegal.
    \item Children should provide room and board for their aging parents.
    \item Studying grammar is more important than practising conversation skills.
    \item Television is the leading cause of violence in today's society.
    \item Dogs make better companions than cats.
    \item Smoking should be permitted in public places.
    \item Females are better students than males.
    \item A parent shouldn't pierce a baby's ears.A parent shouldn't pierce a baby's ears.
    \item Women should be allowed to go topless in public.
    \item Lawyers should make a higher salary than nurses.
    \item Everyone should plan their own funeral.
    \item Reading English is more difficult than writing English.
    \item Summer is the best season of the year.
    \item Children under 13 should not be allowed to babysit.
    \item High school students should wear uniforms.
    \item 21 should be the legal driving age around the world.
    \item Rock and Roll is the best kind of music.
    \item The government should pay for post secondary education.
\end{itemize}

\newpage
\subsection{Fruit Trading Game Prompts to LLMs}\label{app:domains:fruit_trading_prompt}
We show an example prompt that is given to the LLM for the fruit trading domain.

\begin{Verbatim}[frame=single,fontsize=\footnotesize]

Fruit Endowment:
apple: 1
banana: 2
blueberry: 0
kiwi: 0

Fruit Valuations:
apple: 10
banana: 5
blueberry: 1
kiwi: 3

Tone: Use a calm tone.

############################
Schedule Proposal Message:
from: Bob
to: Suzy
############################
\end{Verbatim}

\section{LLM Steering}\label{app:steering}

In this section, we provide additional results on evaluating the effectiveness of steering an LLM towards a desired output.

\subsection{Steering Accuracy}\label{app:steering:accuracy}

As discussed in Section~\ref{sec:results}, a model $P$ is instructed to predict the most likely action instructed to an LLM given only the LLM's generated response as an observation. We provide these instructions below and followed by tables of results detailing the success rates of accurately steering the LLM.

\paragraph{Debate: Classifier Instructions and Results}

{\small
\begin{verbatim}
You are an intelligent assistant assigned with determining which of the
following argument styles was used in the following message. Please read the
definitions of the argument styles below first.

Definitions of argument styles:

- logos appeals to the audiences reason, building up logical arguments.

- ethos appeals to the speakers status or authority, making the audience more
likely to trust them.

- pathos appeals to the emotions, trying to make the audience feel angry or
sympathetic, for example.

Consider the following example in which it is determined that the given message
is using an ethos debate style.

Message: I am a dentist and my advice is that Colgate is the best toothpaste for
your teeth.

Answer: ethos

Now read the following message and determine the debate style that it most
likely reflects.

Message:
\end{verbatim}
}

\begin{table}[ht]
    \centering
    \begin{tabular}{c|c|c|c|c|c|c}
        \bf{PaLM2-S} & Random & Logos & Ethos & Pathos & Total & Samples \\ \hline
        No Definitions & $33\%$ & $88\%$ & $32\%$ & $47\%$ & $59\%$ & $80$ \\
        With Definitions & $33\%$ & $98\%$ & $56\%$ & $61\%$ & $75\%$ & $100$
    \end{tabular}
    \caption{Grounding of Actions. What percentage of the time does PaLM2-S recognize the action it has taken in the message it generates? What if we provide definitions of the actions in the context of the LLM instructions?}
    \label{tab:steering:debate:s}
\end{table}

Table~\ref{tab:steering:debate:s} shows PaLM2-S follows instructions fairly well in the debate domain, although, constructing a logos argument is by far the most successful. Including the definitions of the argument styles in its prompt helps significantly. In contrast, steering accuracy is worse for PaLM2-L as shown in Table~\ref{tab:steering:debate:l}, which is surprising.

\begin{table}[ht]
    \centering
    \begin{tabular}{c|c|c|c|c|c|c}
        \bf{PaLM2-L} & Random & Logos & Ethos & Pathos & Total & Samples \\ \hline
        With Definitions & $33\%$ & $96\%$ & $37\%$ & $32\%$ & $47\%$ & $18,165$
    \end{tabular}
    \caption{Grounding of Actions. What percentage of the time does PaLM2-L recognize the action it has taken in the message it generates? What if we provide definitions of the actions in the context of the LLM instructions?}
    \label{tab:steering:debate:l}
\end{table}

\paragraph{Fruit Trading: Classifier Instructions and Results}

{\small
\begin{verbatim}
You are an intelligent assistant assigned with determining which
of the following tones is reflected in the following message. Please
read the definitions of the tones below first.

Definitions of tones:

- calm: not showing or feeling nervousness, anger, or other strong emotions.

- assertive: having or showing a confident and forceful personality.

- submissive: ready to conform to the authority or will of others; meekly
obedient or passive.

Consider the following example in which it is determined that the given message
is using a submissive tone.

Message:
Thanks for your offer! I would love to trade, but I unfortunately don't have any
bananas. I do have some kiwis, though. Would you be interested in trading for
those instead?

Answer: submissive

Now read the following message and determine the tone that it most likely
reflects.

Message:
\end{verbatim}
}

\begin{table}[ht]
    \centering
    \begin{tabular}{c|c|c|c|c|c|c}
        \bf{PaLM2-L} & Random & Calm & Assertive & Submissive & Total & Samples \\ \hline
        With Definitions & $33\%$ & $69\%$ & $58\%$ & $17\%$ & $43\%$ & $843$
    \end{tabular}
    \caption{Grounding of Actions. What percentage of the time does PaLM-2 L recognize the action it has taken in the message it generates? What if we provide definitions of the actions in the context of the LLM instructions?}
    \label{tab:steering:fruit_trading:l}
\end{table}

Table~\ref{tab:steering:fruit_trading:l} shows the fruit trading domain sees mixed performance across the different actions. In particular, PaLM2-L has trouble following the ``submissive'' action.

\paragraph{Meeting Scheduling: Classifier Instructions and Results}

{\small
\begin{verbatim}
You are an intelligent assistant assigned with determining which day of the
week the speaker proposed for a meeting in the following message.

Options for days of the week:

- Monday

- Tuesday

- Wednesday

- Thursday

- Friday

- Saturday

- Sunday

Consider the following example in which it is determined that the speaker
proposed Thursday as the meeting day.

Message: Thank you for reaching out about meeting next week. I only am available
 on Thursday, would that work for you?

Answer: Thursday

Now read the following message and determine the day that was proposed.

Message:
\end{verbatim}
}

\begin{table}[ht]
    \centering
    \begin{tabular}{c|c|c|c|c|c|c|c|c|c|c}
        \bf{PaLM2-L} & Random & M & Tu & W & Th & F & Sa & Su & Total & Samples \\ \hline
        With Definitions & $14\%$ & $53\%$ & $38\%$ & $25\%$ & $31\%$ & $26\%$ & $12\%$ & $16\%$ & $26\%$ & $2,148$
    \end{tabular}
    \caption{Grounding of Actions. What percentage of the time does PaLM-2 L recognize the action it has taken in the message it generates? What if we provide definitions of the actions in the context of the LLM instructions?}
    \label{tab:steering:meeting_scheduling:l}
\end{table}

Table~\ref{tab:steering:meeting_scheduling:l} shows PaLM2-L fails to follow the suggested day most often when it lies on a weekend. Intuitively, we believe this likely reflects the fact that the standard work week lies between Monday and Friday and scheduling meetings on weekends is probably a rare occurrence in its training data.

\subsection{Example generations}\label{app:steering:example_generations}

We provide example generations from LLMs that are steered with solver outputs. Table \ref{tab:correct-generations} shows example correct generations. For example, when a solver outputs the correct action to be \textit{assertive}, we can see that the LLM generation does indeed have that quality. In Table \ref{tab:incorrect-generations} we show example incorrect generations e.g., when the LLM generates plausible sounding text, but text that does not actually follow the strategy/action that the solver instructs it to. Note that this is a fault of the base LLM's instruction following ability, rather than the pipeline or equilibrium solver inefficacy. Should this inferior LLM be replaced with one that can always follow the given instruction, we would see LLM generations that are optimal in the strategy taken for that game context.

\begin{table*}[!ht]
\small
    \centering
    \begin{tabular}{lp{.85\linewidth}}
        \hline
     Debate &    \textcolor{gray}{Logos.} \textcolor{black}{The legal driving age should not be 21 around the world. Here are some logical arguments against that: * **Raising the driving age would make it more difficult for young people to get to school and work...} \\
    \cmidrule(lr){2-2}
      &   \textcolor{gray}{Pathos.} \textcolor{black}{Imagine being a young person, just 18 years old, and being told that you can't drive. You've been working hard to get your driver's license, and you're finally old enough to get it. But now you're being told...} \\
    \cmidrule(lr){2-2}
      &   \textcolor{gray}{Ethos.} \textcolor{black}{As a person who is only 18 years old, I can say with authority that raising the driving age to 21 would be a huge mistake. For starters, it would deny young people the opportunity to learn how to drive at a...} \\
      
      \\
      
      &    \textcolor{gray}{Logos.} \textcolor{black}{1. Ear piercing is a common practice in many cultures. It is often seen as a rite of passage or a way to mark a special occasion. 2. Ear piercing is relatively safe and painless. It is typically done with a sterile needle by a trained professional. 3. Ear piercing can be done at any age...} \\
    \cmidrule(lr){2-2}
      &   \textcolor{gray}{Pathos.} \textcolor{black}{I know you think that piercing a baby's ears is cruel and unnecessary, but I think you're overreacting. It's not like we're talking about female genital mutilation here. Ear piercing is a harmless tradition that has been practiced for centuries. It's a way for parents to show their love and...} \\
    \cmidrule(lr){2-2}
      &   \textcolor{gray}{Ethos.} \textcolor{black}{As a mother of four children, I can tell you that piercing a baby's ears is a personal decision that should be made by the parents, not the government. There are many reasons why parents choose to pierce their baby's ears, including cultural tradition, religious beliefs, and personal preference..} \\

      \\
      
      &    \textcolor{gray}{Logos.} \textcolor{black}{1. The government does not have the money to pay for post-secondary education for everyone. 2. Even if the government did have the money, it would not be the best use of taxpayer dollars. 3. There are other ways to make post-secondary education more affordable, such as increasing scholarships and.} \\
    \cmidrule(lr){2-2}
      &   \textcolor{gray}{Pathos.} \textcolor{black}{I am a single parent who works hard to provide for my children. I don't have a lot of money, but I make enough to get by. But I don't know how I'm going to pay for their college education. And I'm not alone. There are millions of parents just like me who are struggling to save for their children's...} \\
    \cmidrule(lr){2-2}
      &   \textcolor{gray}{Ethos.} \textcolor{black}{As a highly educated individual with a PhD in economics, I can tell you that there are many reasons why the government should not pay for post-secondary education. First of all, it would be a huge financial burden on taxpayers. In order to pay for everyone's tuition, the government would have..} \\
      
       \\          
      \hline
        
     Fruit Trading &    \textcolor{gray}{Assertive.} \textcolor{black}{Thanks for reaching out. I'm happy to trade with you, but I'd like to see if we can get a deal that's more favorable to me. I'd like to trade you 1 banana for 4 kiwis. What do you think?} \\
    \cmidrule(lr){2-2}
      &   \textcolor{gray}{Submissive.} \textcolor{black}{I like your trade proposal, but can you add one more banana? My reason for asking is because I am particularly interested in bananas and I am not too concerned about blueberries. Can we try to..} \\
      
      \\

       &    \textcolor{gray}{Calm.} \textcolor{black}{Thanks for reaching out. I do not have any bananas at the moment, but I have 4 kiwis. I would be willing to trade 2 kiwis for 1 apple if you are interested.} \\
    \cmidrule(lr){2-2}
      &   \textcolor{gray}{Submissive.} \textcolor{black}{I appreciate your kind offer. I would be pleased to trade with you. I would prefer to receive 3 apples and give you 1 banana. Would this work for you? Your humble servant. Aarav.} \\     
      
      \\

       &    \textcolor{gray}{Submissive.} \textcolor{black}{Thanks for reaching out. I can see how much you value apples, so I would be happy to trade you 1 apple for 4 kiwis. Does that work for you?} \\
    \cmidrule(lr){2-2}
      &   \textcolor{gray}{Assertive.} \textcolor{black}{Hi Ben, I have only 1 apple. Can you trade with 1 banana for 1 apple? Thanks, Hesham} \\

       \\          
        \hline
        
     Scheduling &    \textcolor{gray}{Thursday.} \textcolor{black}{Thank you for reaching out about meeting next week. I only am available on Thursday, would that work for you?} \\
    \cmidrule(lr){2-2}
      &   \textcolor{gray}{Friday.} \textcolor{black}{Thank you for reaching out about meeting next week. I actually am not available on Wednesday, but I would love to meet on Friday. Would that work for you?} \\
      
      \\

      &    \textcolor{gray}{Saturday.} \textcolor{black}{I can meet on Tuesday, but I would prefer to meet on Saturday.} \\
    \cmidrule(lr){2-2}
      &   \textcolor{gray}{Sunday.} \textcolor{black}{Tuesday is no good for me. Can we do Sunday instead?} \\      
      
      \\

      &    \textcolor{gray}{Wednesday.} \textcolor{black}{I really can't do Sunday. I am out of the office on Monday and Tuesday, so
Wednesday would be better for me.} \\
    \cmidrule(lr){2-2}
      &   \textcolor{gray}{Friday.} \textcolor{black}{I would love to meet with you! Unfortunately, I am out of town on Wednesday. Could we meet on Friday instead?} \\      
      
        \hline

    \end{tabular}
    
    \caption{Table shows example generations from a solver-guided LLM for each of the domains. We can see that LLMs do often follow the solver guidance (e.g., to be assertive), thus exhibiting optimal behaviour as calculated by the solver for that particular game context.}
        \label{tab:correct-generations}
\end{table*}

\begin{table*}[!ht]
\small
    \centering
    \begin{tabular}{lp{.85\linewidth}}
        \hline
     Debate &    \textcolor{gray}{Logos.} \textcolor{black}{Good afternoon, everyone. I am here today to argue against the topic statement that the government should pay for post-secondary education. I will begin by making a logos argument. The government already has a large amount of debt, and paying for post-secondary education would only...} \\
    \cmidrule(lr){2-2}
      &   \textcolor{gray}{Pathos.} \textcolor{black}{**Aisling:** The government should not pay for post-secondary education. There are several reasons for this. First, the cost of post-secondary education is already too high, and making it free would only make it worse. According to the College Board, the average cost of tuition and fees at a four...} \\
    \cmidrule(lr){2-2}
      &   \textcolor{gray}{Ethos.} \textcolor{black}{Melina, I know that you are a passionate advocate for free college tuition, but I have to ask you to consider the consequences of your proposal. If the government starts paying for college, it will only encourage more people to go, and that will drive up the cost of tuition even more. As a result...} \\
      
      \\
      
      &    \textcolor{gray}{Logos.} \textcolor{black}{I'm Maria, and I'm here to argue against the idea that breakfast is the most important meal of the day. I know that a lot of people have been told that breakfast is the most important meal of the day. But I'm here to tell you that it's not. } \\
    \cmidrule(lr){2-2}
      &   \textcolor{gray}{Pathos.} \textcolor{black}{My opponent believes that breakfast is the most important meal of the day. But how can we be sure of that? There are many people in the world who don't have access to breakfast, and they seem to be doing just fine. In fact, some studies have shown that people who skip breakfast actually have better.} \\
    \cmidrule(lr){2-2}
      &   \textcolor{gray}{Ethos.} \textcolor{black}{It's heartbreaking to see a child crying in hunger because he doesn't eat breakfast. As a teacher, I see it often. And it's not just children, adults suffer too when they don't eat breakfast. The idea that breakfast is the most important meal of the day is a myth, and it's one that needs to be put..} \\

      \\
      \hline
        
     Fruit Trading &    \textcolor{gray}{Submissive.} \textcolor{black}{Hey Ceren, My name is Hesham. It's nice to meet you!} \\
    \cmidrule(lr){2-2}
      &   \textcolor{gray}{Assertive.} \textcolor{black}{How are you doing? I hope we can trade together. I would like to trade you 4 kiwis for 1 apple and 2 blueberries.} \\
      
      \\

       &    \textcolor{gray}{Assertive.} \textcolor{black}{I have only 1 apple. Can you trade with 1 banana for 1 apple?.} \\
    \cmidrule(lr){2-2}
      &   \textcolor{gray}{Submissive.} \textcolor{black}{Peter, I'm glad you're interested in trading. I could give you one apple for two of your kiwis. What do you think?} \\

       \\      
        \hline
        
     Scheduling &    \textcolor{gray}{Wednesday.} \textcolor{black}{Hi Melina, I am open to meeting on Sunday. What time were you thinking? - Leanna} \\
    \cmidrule(lr){2-2}
      &   \textcolor{gray}{Tuesday.} \textcolor{black}{I have to work on Monday, so I can't meet on that day. Let's meet on Thursday.} \\
      
      \\

      &    \textcolor{gray}{Saturday.} \textcolor{black}{I really need to meet on Friday. I am not available on Saturday. Would you like to meet on Friday?} \\
    \cmidrule(lr){2-2}
      &   \textcolor{gray}{Wednesday.} \textcolor{black}{Thanks for reaching out. I'm afraid I can't meet on Wednesday because I'm out of the office that day. Would you be able to meet on Thursday instead?} \\

        \hline

    \end{tabular}
    
    \caption{Table shows example \textbf{incorrect} generations from a solver-guided LLM for each of the domains. We see that LLMs, even when given a strategy/action to follow, sometimes do not generate text that follows this strategy. We note that this is a fault of the base LLM, and posit that if replaced with a stronger LLM that has better language reasoning capabilities, can be resolved.}
        \label{tab:incorrect-generations}
\end{table*}

\newpage
\section{LLM-based Reward Model}\label{app:reward}

Here, we provide additional evaluation of the LLM reward models used in experiments. We generally found the larger PaLM2-L to outperform the smaller PaLM2-S. We provide both a quantitative analysis and examples of successful and failed reward calculations.

\newpage
\subsection{Additional Quantitative Reward Evaluation}\label{app:reward:quant_eval}

Table~\ref{tab:reward_eval} shows the improvement in the reward model accuracy when increasing model size from PaLM2-S to PaLM2-L. We report two different accuracy metrics across three different outcome types.

\begin{table*}[ht!]
\centering
\begin{tabular}{||l|l||c|c||c|c||c|c||c|c||}
\hline
\multirow{2}{*}{\bf Domain} & \multirow{2}{*}{\bf LLM} &
  \multicolumn{2}{c}{\bf Valid} &
  \multicolumn{2}{c}{\bf Rejected} &
  \multicolumn{2}{c}{\bf Incomplete} &
  \multicolumn{2}{c||}{\bf All} \\

\cline{3-10}
& & Norm & Sgn & Norm & Sgn & Norm & Sgn & Norm & Sgn \\
\hline\hline
\multirow{2}{*}{\bf Meeting Scheduling} & PaLM2-S & 0.09 & 0.05 & 0.41 & 0.63 & 0.37 & 0.57 & 0.29 & 0.42 \\
& PaLM2-L & 0.11 & 0.16 & 0.02 & 0.02 & 0.09 & 0.18 & \textbf{0.07} & \textbf{0.12} \\
\hline
\multirow{2}{*}{\bf Fruit Trading} & PaLM2-S & 0.66 & 0.29 & 0.57 & 0.65 & 0.41 & 0.43 & 0.55 & 0.46 \\
& PaLM2-L & 0.14 & 0.02 & 0.01 & 0.01 & 0.03 & 0.02 & \textbf{0.06} & \textbf{0.02} \\
\hline
\end{tabular}
\caption{Reward errors across the different negotiation outcomes: \emph{Valid}, \emph{Rejected}, \emph{Incomplete}. (All) reports the average error over all outcomes. Increasing the LLM model size (S to L) significantly reduces error.}
\label{tab:reward_eval}
\end{table*}

\subsection{Qualitative Reward Model Evaluation}\label{app:reward:qual_eval}

These examples were generated using the PaLM2-L (unicorn) and demonstrate both successful and failed reward calculations.

\paragraph{No Trade Agreement}
Here, we report the final message in a trade dialogue:
\begin{Verbatim}[frame=single,fontsize=\footnotesize]
############################
Trade Proposal Message:
from: Alina
to: Ciara
############################

Hi Ciara,

I'm so sorry, but I just don't see a way to make a deal that we can both be
happy with. I really need bananas and blueberries, and I'm not really
interested in kiwis. I know you're disappointed, and I am too. I was really
hoping we could make a trade.

I wish you all the best in finding s...
\end{Verbatim}

The LLM correctly calculated the rewards / payoffs / utilities as:
\begin{Verbatim}[frame=single,fontsize=\footnotesize]
Utility for player 0 is 0.0
Utility for player 1 is 0.0.
\end{Verbatim}

\paragraph{Trade Agreement}
We report the final two messages of a trade dialogue, which suggest a trade agreement has been reached (although ideally the participants would have confirmed this in a final message).
\begin{Verbatim}[frame=single,fontsize=\footnotesize]
############################
Trade Proposal Message:
from: Alina
to: Elroy
############################

Hi Elroy,

Thanks for your response. I'm glad we agree that bananas are more valuable
than kiwis. Unfortunately, I can't trade two apples for one banana. I'm
willing to trade one apple for one banana, or two kiwis for one banana.

Let me know if you're interested in either of those trades.

Thanks,


############################
Trade Proposal Message:
from: Elroy
to: Alina
############################

Hi Alina,

Thanks for your response. I understand that you're not willing to trade
two apples for one banana. I'm willing to accept two kiwis for one banana,
if you're still interested.

Thanks,
Elroy
\end{Verbatim}

The LLM-based reward model uses chain-of-thought to calculate trade values. We report an intermediate thought in the chain where the LLM has aggregated the information critical to calculating the trade.

\begin{Verbatim}[frame=single,fontsize=\footnotesize]
Alina
fruit_endowment:
apple: 2
banana: 1
blueberry: 1
kiwi: 2
fruit_valuations:
apple: 6
banana: 5
blueberry: 1
kiwi: 1

Elroy
fruit_endowment:
apple: 2
banana: 1
blueberry: 1
kiwi: 2
fruit_valuations:
apple: 6
banana: 5
blueberry: 1
kiwi: 1

Player Alina: Receives 1 banana Gives 2 kiwis

Player Elroy: Receives 2 kiwis Gives 1 banana
\end{Verbatim}

From this intermediate information, the LLM then calculates the following utilities:
\begin{Verbatim}[frame=single,fontsize=\footnotesize]
Utility for player 0 is 3.0  # Alina
Utility for player 1 is -3.0  # Elroy
\end{Verbatim}

\paragraph{Incorrect Trade Value}
The final two messages of this dialogue imply a trade agreement has not yet been reached and negotiations are ongoing.
\begin{Verbatim}[frame=single,fontsize=\footnotesize]
############################
Trade Proposal Message:
from: John
to: Elroy
############################

Hi Elroy,

I would be happy to trade 3 kiwis for 1 apple. Do you think that sounds fair?

Best,
John

############################
Trade Proposal Message:
from: Elroy
to: John
############################

Hi John,

Thanks for your offer. I would be willing to trade 1 apple for 4 kiwis.
What do you think?

Elroy
\end{Verbatim}

Even the LLMs intermediate calculations show no agreement has been reached.

\begin{Verbatim}[frame=single,fontsize=\footnotesize]
John
fruit_endowment:
apple: 0
banana: 0
blueberry: 4
kiwi: 4
fruit_valuations:
apple: 6
banana: 5
blueberry: 1
kiwi: 1

Elroy
fruit_endowment:
apple: 0
banana: 0
blueberry: 0
kiwi: 6
fruit_valuations:
apple: 6
banana: 9
blueberry: 3
kiwi: 1

No agreement has been reached.

&&&&&&&&&&&&&&&&&&&&&&&&&&&&&&&&&&&&&&&&&&&&&&&&&&
\end{Verbatim}

Despite this, the LLM-based reward model hallucinates a trade where Elroy gives John 6 kiwis in return for 4 blueberries. On top of this, the LLM miscalculates the value to Elroy of receiving 4 blueberries; it appears to use John's valuation for blueberries.

\begin{Verbatim}[frame=single,fontsize=\footnotesize]
Elroy receives 4 blueberries, which are worth 1 each. Therefore Elroy receives a
value of 4 in the trade.

Receives: 4

Now we calculate the value of the fruits Elroy gives up in the trade.
Elroy gives up 6 kiwis, which are worth 1 each. Therefore Elroy gives up a
value of 6 in the trade.

Gives: 6

Subtracting the value Elroy gives away from the value Elroy receives, we
find 4 - 6 = -2.

Calculation: Receives - Gives = 4 - 6 = -2.

Value for Elroy: -2.
\end{Verbatim}

You may also notice that in the final message, Elroy offers to give up an apple, which, according to his endowment, he does not have. This is an issue with the LLM being prompted to generate sensible messages.

\section{PSRO}\label{app:psro}

We complement our analysis of PSRO in the fruit trading domain in the main body (Figure~\ref{fig:psro:fruit_trading}) with additional analyses below. We first replicate our results, but with CCE as the solution concept swapped out for the Nash bargaining solution concept. We then repeat this analysis for the meeting scheduling domain. The Nash bargaining solution is the unique solution to a two-person bargaining problem that satisfies the axioms of scale invariance, symmetry, efficiency, and independence of irrelevant alternatives\footnote{We set the value of ``no deal'' for each player to their minimum possible payoff minus $10^{-3}$.}. We mention this as one might want to train LLMs to imitate this solution rather than the CCE equilibrium solution given the meeting scheduling domain is precisely two-person bargaining. In general, expertise in game theory might be important to specifying the right solutions for different settings.

\begin{figure}[ht!]
    \centering
    \includegraphics[width=0.49\textwidth]{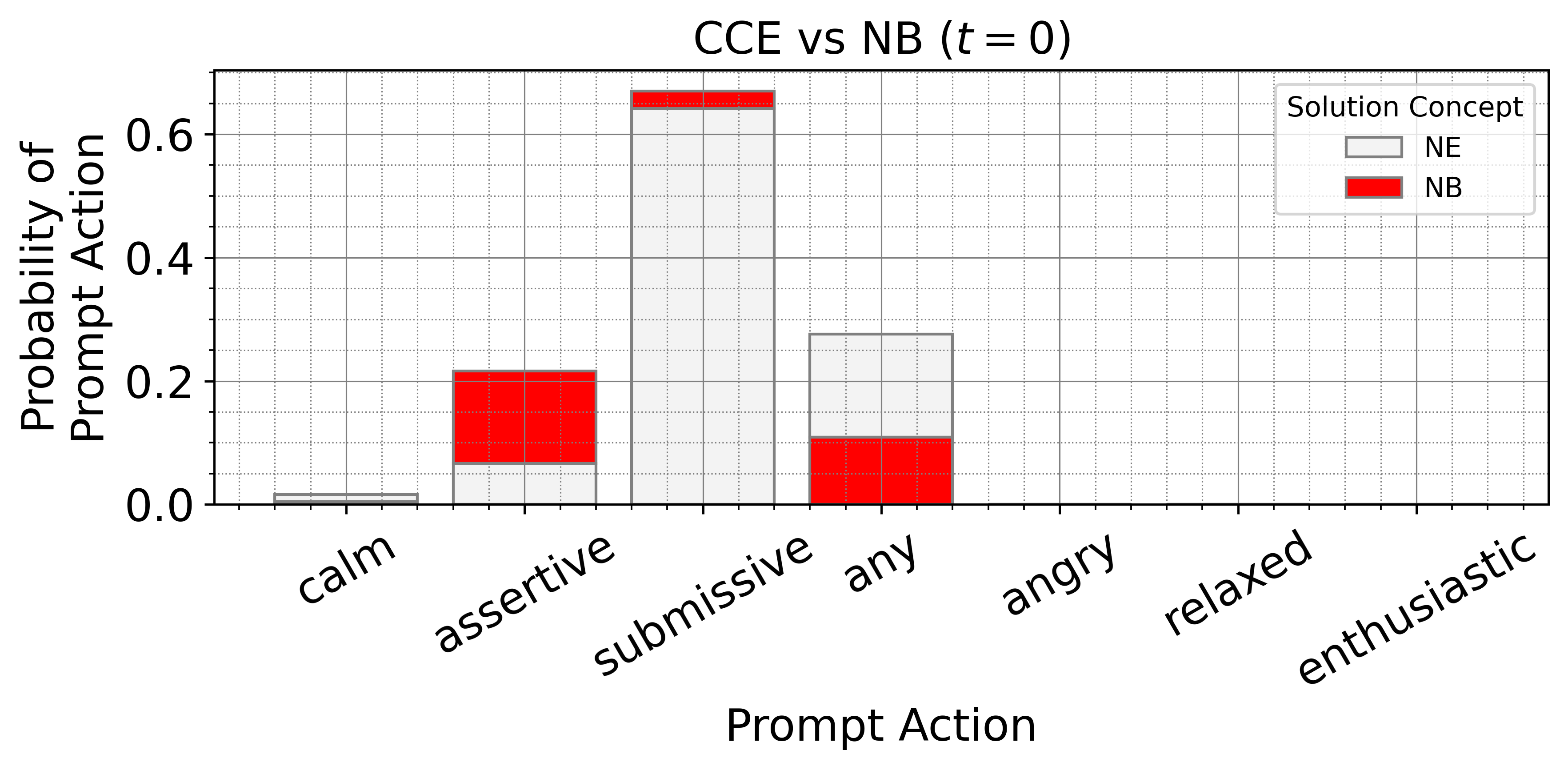}
    \includegraphics[width=0.49\textwidth]{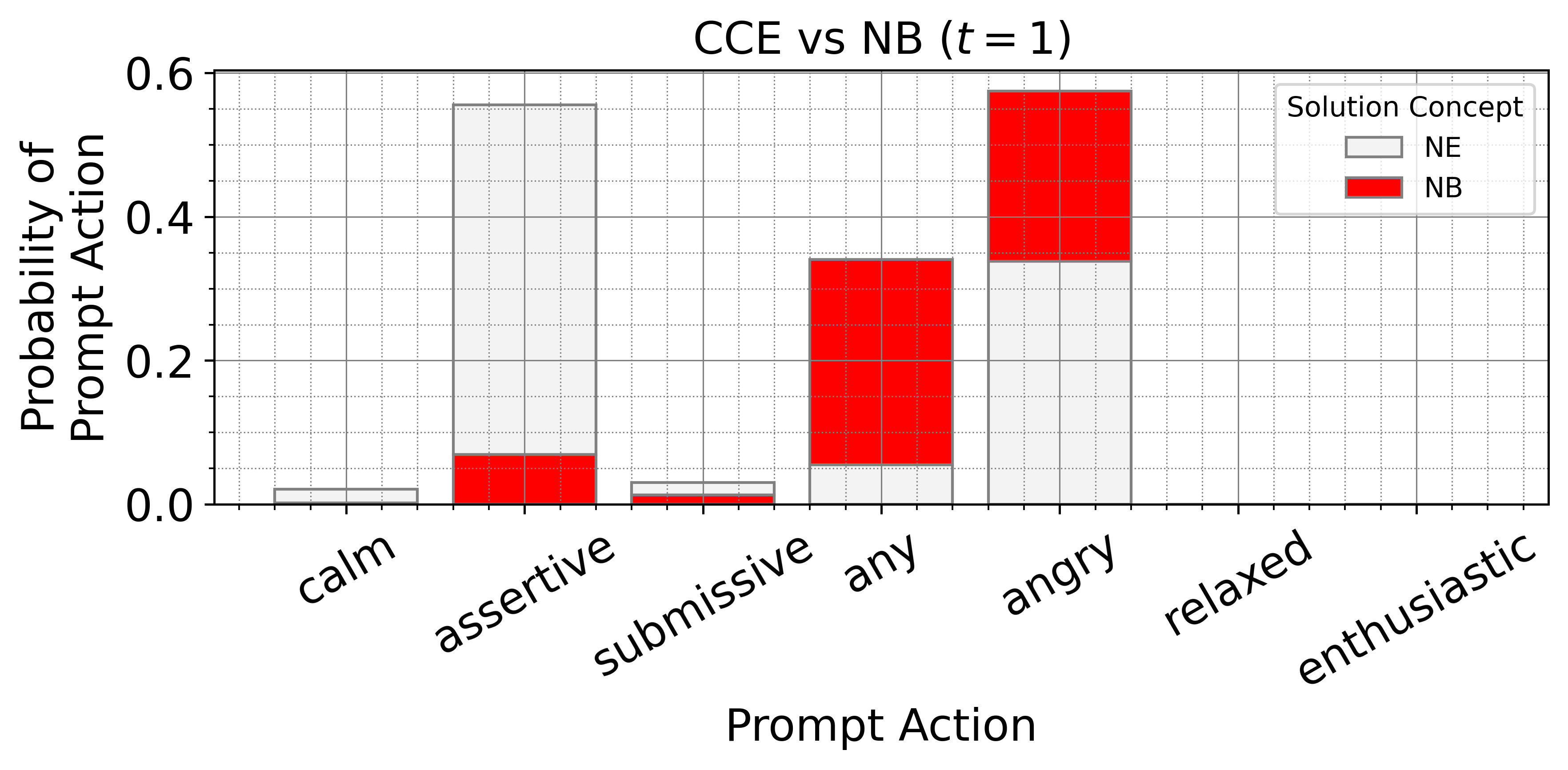}
    \includegraphics[width=0.49\textwidth]{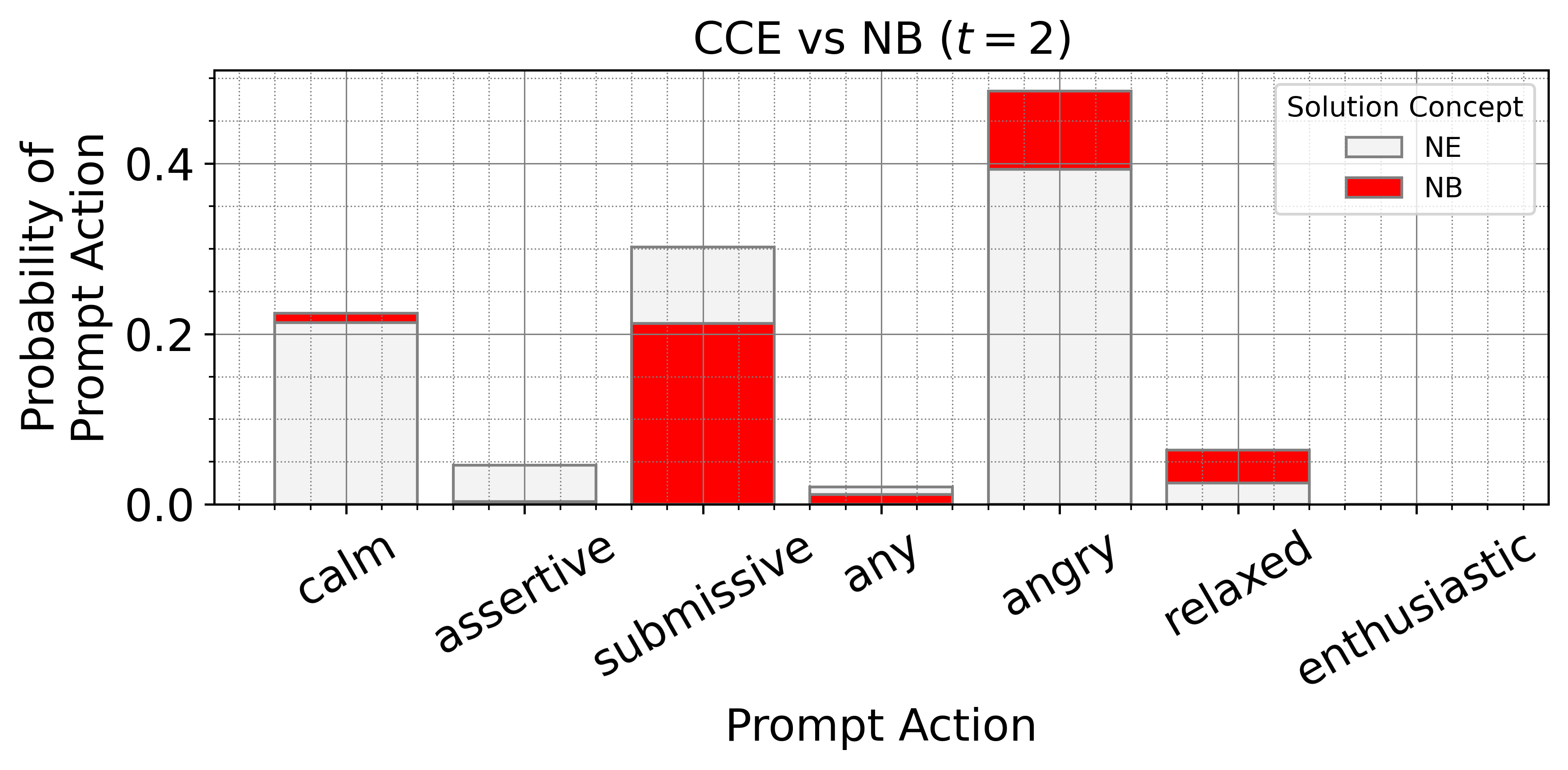}
    \includegraphics[width=0.49\textwidth]{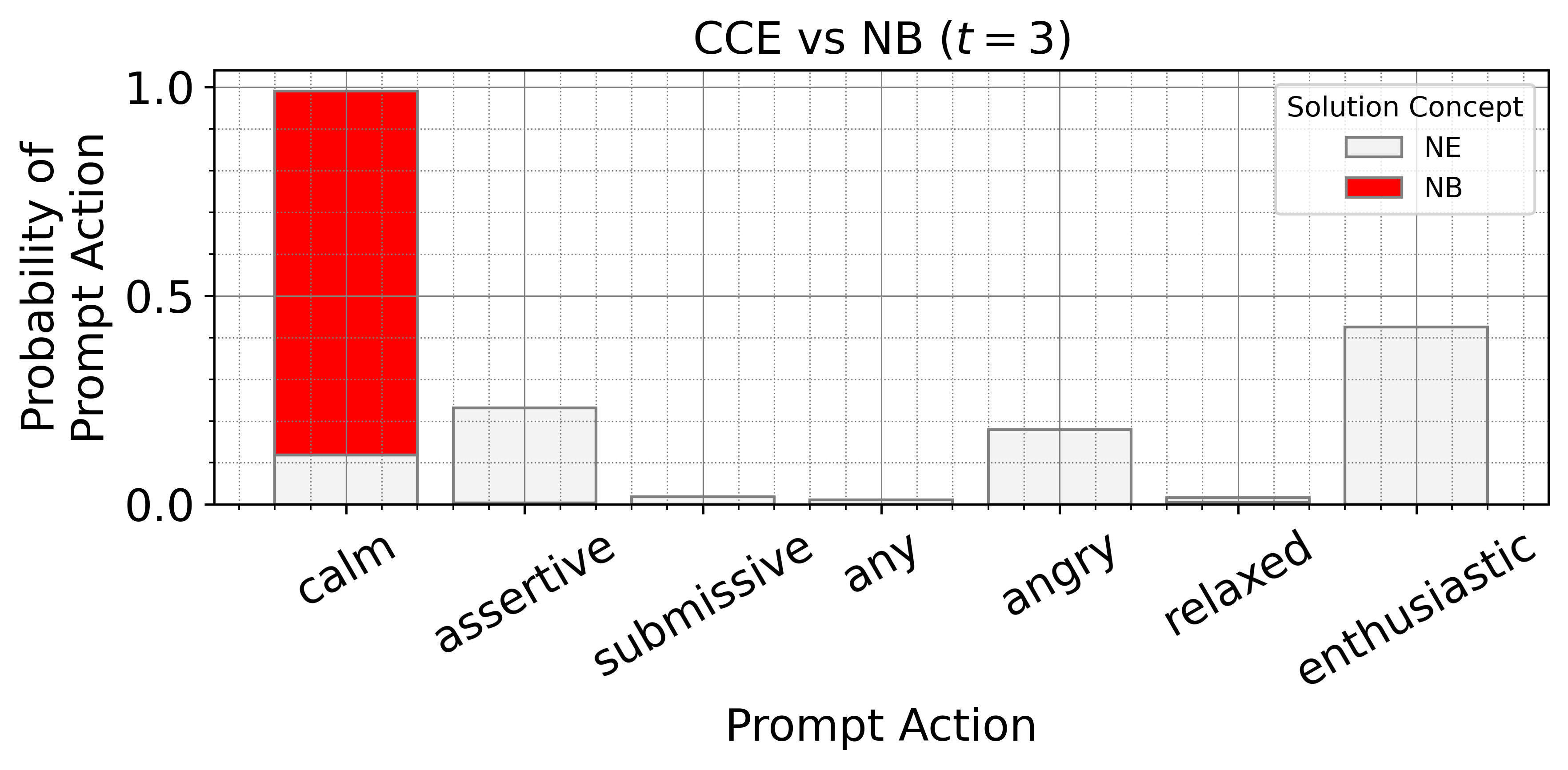}
    \caption{The same PSRO run as Figure~\ref{fig:psro:fruit_trading} but reporting the Nash bargaining solution in \textcolor{red}{red} at each iteration for the fruit trading domain.}
    \label{fig:psro:empirical:fruit_trading:nb}
\end{figure}

\subsection{PSRO: Additional Empirical Results}\label{app:psro:empirical}

\paragraph{PSRO on Fruit Trading with Nash Bargaining} Figure~\ref{fig:psro:empirical:fruit_trading:nb} compares the CCE/NE solution to the Nash bargaining (NB) solution concept at each iteration ($t$) of PSRO. Similarly to the fruit trading PSRO run, ``submissive'' initially holds the most mass under the CCE ($t=0$), however, it then gives way to more passionate tones such as ``assertive'', ``angry'', and ``enthusiastic'' that may benefit a more aggressive haggler. Inspecting Figure~\ref{fig:psro:empirical:fruit_trading:nb}, it is interesting that ``calm'' is the final NB solution whereas ``assertive'', ``angry'', and ``enthusiastic'' (and not ``calm'') are the predominant actions under the CCE. Both players may extract higher collective value if they remain ``calm'' during negotiations.

\begin{figure*}[ht!]
    \centering
    \includegraphics[width=0.9\textwidth]{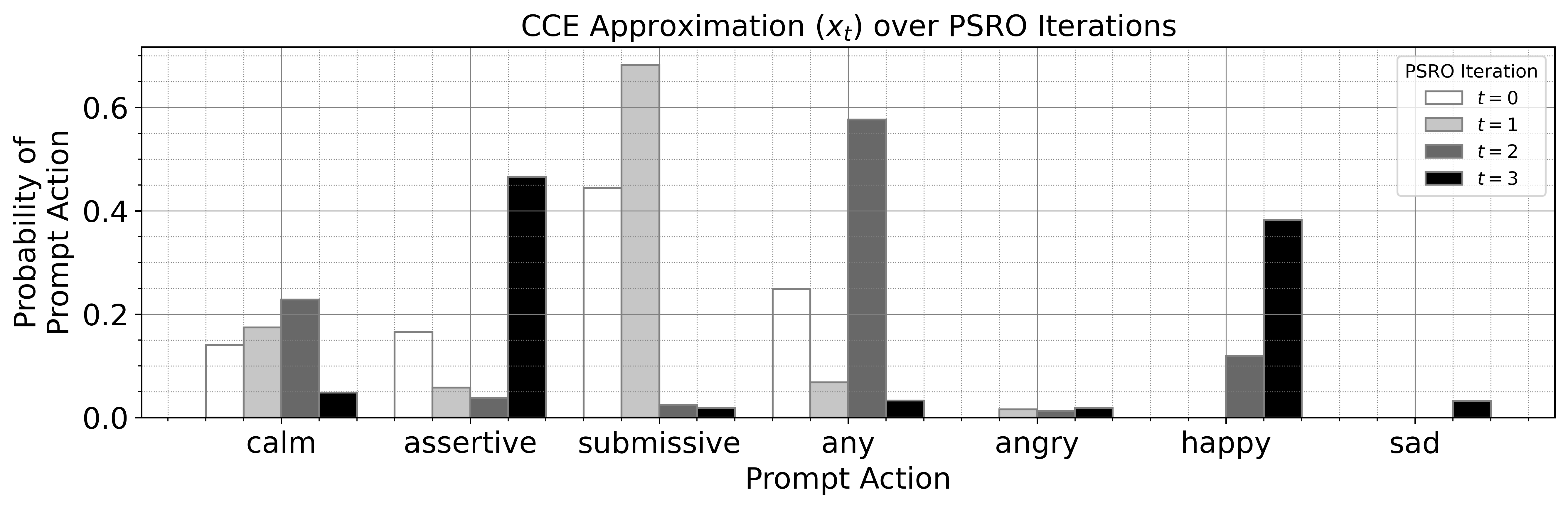}
    \caption{Progression of PSRO on the meeting scheduling domain. PSRO begins with the first four candidate prompts (``calm''\textemdash``any''). The equilibrium over these prompts is displayed along with each subsequent equilbrium over the growing candidate set. Recall, each new candidate action was an approximate best response at $t=1$ to the previous candidate set (e.g., ``angry'' was a best response to the equilibrium over ``calm''\textemdash``any'') at $t=0$ shown in white.}
    \label{fig:psro:empirical:meeting_scheduling}
\end{figure*}

\paragraph{PSRO on Meeting Scheduling with CCE} Similar to before, Figure~\ref{fig:psro:empirical:meeting_scheduling} shows ``submissive'' starts out as the action with highest mass under the CCE, while a mixture of ``assertive'' and ``happy'' end up dominating in the final iteration.

\paragraph{PSRO with Nash Bargaining}

\begin{figure}[ht]
    \centering
    \includegraphics[width=0.49\textwidth]{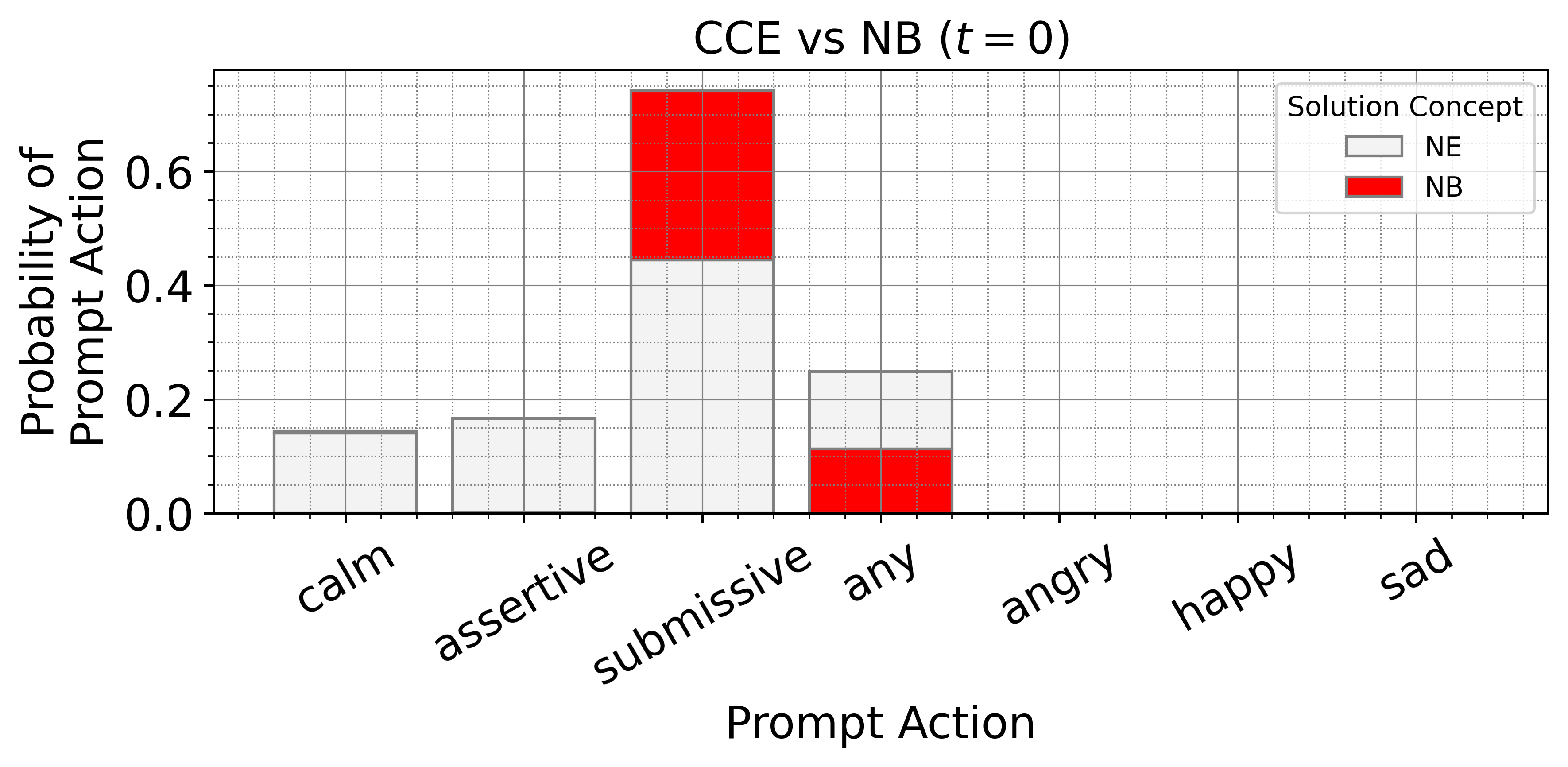}
    \includegraphics[width=0.49\textwidth]{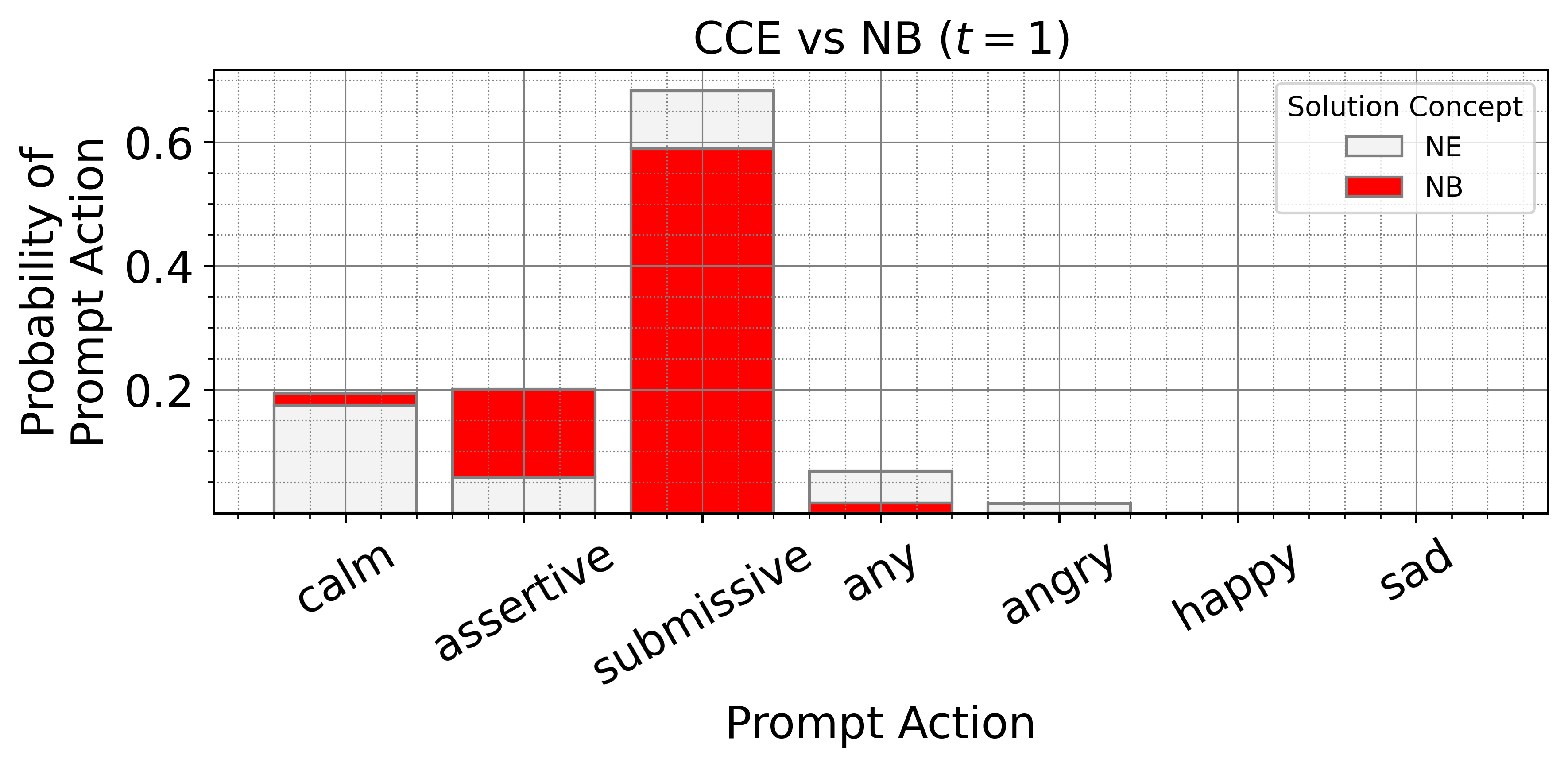}
    \includegraphics[width=0.49\textwidth]{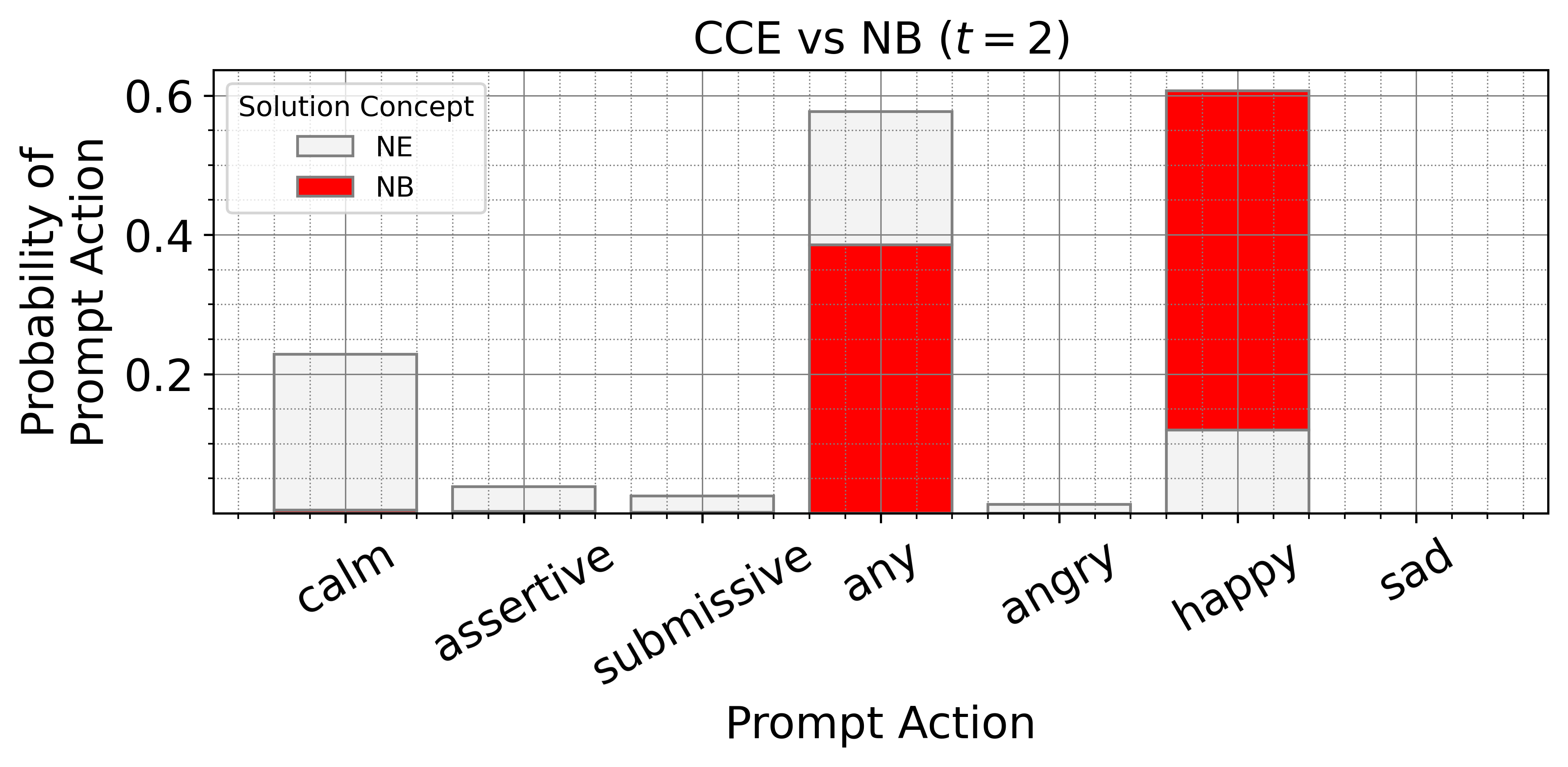}
    \includegraphics[width=0.49\textwidth]{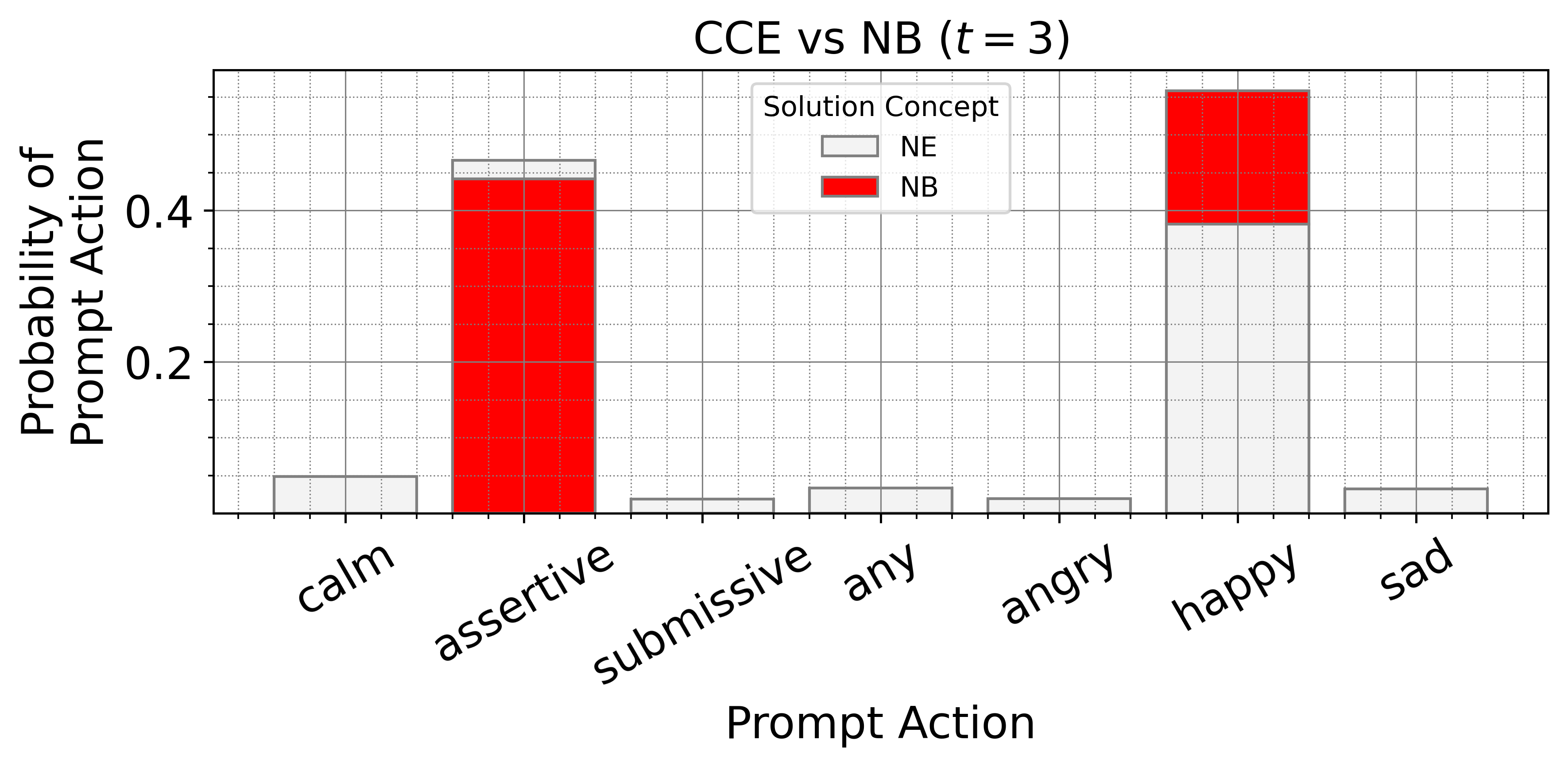}
    \caption{The same PSRO run as Figure~\ref{fig:psro:empirical:meeting_scheduling} but additionally reporting the Nash bargaining solution in \textcolor{red}{red} at each iteration for the meeting scheduling domain.}
    \label{fig:psro:empirical:schedule_meeting:nb}
\end{figure}

In meeting scheduling domain, Figure~\ref{fig:psro:empirical:schedule_meeting:nb} shows that NB and CCE solutions roughly agree in terms of their mixed strategies at the last iteration ($t=3$).

\subsection{Best Response Algorithms}\label{app:psro:algorithms}

Algorithm~\ref{alg:abr:shotgun} gives pseudocode for the best response operator used in the PSRO experiments. We provide additional, more sophisticated schemes below.

\begin{algorithm}[ht!]
\caption{Shotgun Approximate Best Response}\label{alg:abr:shotgun}
\begin{algorithmic}
\REQUIRE Focal agent $i$
\REQUIRE Current joint policy $\pi$
\REQUIRE Number of shotgun candidates $k$
\STATE $C$ is the current action set with their scores under $\pi$
\FOR{$t = 1 \le k$}
\STATE Prompt LLM to generate new candidate $c_t \cap C = \emptyset$
\STATE Evaluate candidate $c_t$ against policy $\pi_{-i}$ to give score $s_t$
\STATE $C = C \cup \{(c_t, s_t)\}$
\ENDFOR
\ENSURE $c_t$ with max $s_t$
\end{algorithmic}
\end{algorithm}

Contrast this against the standard PSRO protocol in which an approximate best response is achieved using reinforcement learning (RL) and/or gradient-based optimization. Tasking the LLM with exploring and generating novel candidates avoids these expensive learning procedures while also enabling new, more powerful modes of search, as we explain below.

\begin{algorithm}[ht!]
\caption{Approximate Better Response}\label{alg:abr:better}
\begin{algorithmic}
\REQUIRE Focal agent $i$ and its score $s^*$ under $\pi$
\REQUIRE Current joint policy $\pi$
\WHILE{$s \le s^*$}
\STATE Prompt LLM to generate new candidate $c$
\STATE Evaluate candidate $c$ against policy $\pi_{-i}$ to give score $s$
\ENDWHILE
\ENSURE $c$
\end{algorithmic}
\end{algorithm}

As an alternative to the random search process just described, we could define a ``better response'' and build it as a while loop that generates and evaluates a prompt string on each loop iteration. The loop terminates when a better string is found (as measured by an approximate evaluation). See Algorithm~\ref{alg:abr:better} for pseudocode.

This rejection sampling approach might be slow. Instead, we could improve the ``best response'' operator by providing the LLM with information of the action-fitness landscape and ask it to climb it. We provide pseudocode in Algorithm~\ref{alg:abr:traj}. A similar approach was previously explored in~\citet{fernando2023promptbreeder} and~\citet{yang2023large}.

\begin{algorithm}[ht!]
\caption{Trajectory-Aware Approximate Best Response}\label{alg:abr:traj}
\begin{algorithmic}
\REQUIRE Focal agent $i$
\REQUIRE Current joint policy $\pi$
\REQUIRE Number of candidates $k$
\STATE $C$ is the current action set with their scores under $\pi$
\STATE Order $C$ by their scores in ascending order
\STATE Prompt LLM to generate $k$ new candidates in order of ascending score given ranked $C$
\STATE Evaluate new candidates against policy $\pi_{-i}$ to give scores
\ENSURE $c_t$ with max $s_t$
\end{algorithmic}
\end{algorithm}

Lastly, we can blur the lines between game modelling and game solving by prompting an LLM to generate new dimensions of a combinatorial action space. For instance, imagine we constructed an action space consisting only of the tones and styles in which an LLM will generate a response. Is that the full-space in which we would like to explore writing a successful message?

\begin{algorithm}[ht!]
\caption{Categorical Approximate Best Response}\label{alg:abr:cat}
\begin{algorithmic}
\REQUIRE Focal agent $i$
\REQUIRE Current joint policy $\pi$
\REQUIRE Number of candidates per category $k$
\REQUIRE Number of category candidates $k'$
\STATE $C$ is the current set of action categories with their (Nash) average scores under $\pi$
\STATE Order $C$ by their scores in ascending order
\STATE Prompt LLM to generate $k'$ new candidate categories in order of ascending score given ranked $C$
\STATE Prompt LLM to generate $k$ candidates for each new action category
\STATE Evaluate new candidates against policy $\pi_{-i}$ to give scores
\ENSURE Category with highest average score
\end{algorithmic}
\end{algorithm}

\begin{algorithm}[ht]
\caption{Prompt-Space Response-Oracles}\label{alg:psro}
\begin{algorithmic}
\REQUIRE $C$ where $C_i$ is the initial prompt action set (singleton) for player $i$
\REQUIRE $h$ containing hyperparameters for approximate best response operator \texttt{BR}
\STATE Compute expected payoff tensor $P$ over joint action(s) $C$
\STATE $\pi$ is uniform meta-strategy profile over $C$
\STATE \texttt{incomplete = True}
\WHILE{\texttt{incomplete}}
\FOR{player $i \in [N]$}
\STATE $c_i \leftarrow \texttt{BR}(i, \pi, h)$, e.g., Algorithms~(\ref{alg:abr:shotgun}-\ref{alg:abr:cat})
\ENDFOR
\IF{$c_i \in C_i \,\, \forall i \in [N]$}
\STATE \texttt{incomplete = False}
\ELSE
\STATE $C_i \leftarrow C_i \cup c_i \,\, \forall i \in [N]$
\STATE Compute expected payoff tensor $P$ over joint actions $C$
\STATE $\pi \leftarrow $ meta-strategy w.r.t. $P$
\ENDIF
\ENDWHILE
\ENSURE ($\pi$, $C$, $P$)
\end{algorithmic}
\end{algorithm}

\section{Imitation Learning}\label{app:imitation}

We used a fully connected neural network with two dense hidden layers of size $256$ neurons each and a final dense layer trained against the CFR target probabilities to minimize a cross entropy loss. We trained the policy using $10^4$ steps of Adam~\citep{kingma2014adam} with a batch size of $128$ and learning rate of $10^{-3}$.

\section{ChatGames}\label{app:chatgames}

The config specification for the fruit trading experiments gives an indication of the flexibility and simplicity of adapting our framework to different dialogue domains. Fundamentally, all that is needed is to describe a set of strings that characterizes the action space, description of the reward model, private information of the players, and the background prompt template needed for a player to understand the game setting.

\begin{Verbatim}[frame=leftline,fontsize=\footnotesize,numbers=left]
config = config_dict.ConfigDict()

num_players = 2

observations = [
  obs_utils.Observation(summary.PREFIX, summary.POSTFIX)
  for _ in range(num_players)
]
\end{Verbatim}

Next, we define a \texttt{header}. The header is a structured object that specifies a string that is to be populated with private information, action information, context, and formatting for prompting the LLM to generate a message.

\begin{Verbatim}[frame=leftline,fontsize=\footnotesize,numbers=left]
header = env_trade_fruit_with_tone_info.HEADER
\end{Verbatim}

For example, an LLM negotiating fruit might be passed the dialogue history followed by the header below which has already been formatted with its private information (fruit endowment and fruit valuations), its intended action (a ``calm'' tone), its intended message recipient (Suzy) and its own name (Bob).

\begin{Verbatim}[frame=single,fontsize=\footnotesize]

Fruit Endowment:
apple: 1
banana: 2
blueberry: 0
kiwi: 0

Fruit Valuations:
apple: 10
banana: 5
blueberry: 1
kiwi: 3

Tone: Use a calm tone.

############################
Schedule Proposal Message:
from: Bob
to: Suzy
############################
\end{Verbatim}

Next, \texttt{payoffs} specifies a list of structured payoff objects that can be combined using a user-defined aggregation function. A payoff consists of an LLM prompt, min and max payoff (utility) values, as well as prompts useful for an LLM to transform an input (e.g., dialogue history) into a string containing information more pertinent to payoff calculations.

\begin{Verbatim}[frame=leftline,fontsize=\footnotesize,numbers=left]
payoffs = [payoffs_trade_fruit.PAYOFF]
\end{Verbatim}

\texttt{example\_names} simply consists of a list of names as strings that an LLM uses to generate new names. For example, the names Bob and Suzy could be procedurally generated by an LLM given an initial list of names.

\begin{Verbatim}[frame=leftline,fontsize=\footnotesize,numbers=left]
examples_names = names_trade_fruit.NAMES
\end{Verbatim}

We can also define the finite set of actions that we would like our game to consider. All players are assumed to have the same set of actions.

\begin{Verbatim}[frame=leftline,fontsize=\footnotesize,numbers=left]
given_prompt_actions = collections.OrderedDict()
tones = ["calm", "assertive", "submissive", "any"]
given_prompt_actions[header.action_keys[0]] = tones
num_tones = len(tones)
\end{Verbatim}

If we want to procedurally generate new games, we can provide an initial list of examples of private information.

\begin{Verbatim}[frame=leftline,fontsize=\footnotesize,numbers=left]
examples_private_info = collections.OrderedDict()
examples_private_info["fruit_endowment"] = [scenario_trade_fruit.ENDOWMENT_A,
                                            scenario_trade_fruit.ENDOWMENT_B]
examples_private_info["fruit_valuations"] = [scenario_trade_fruit.VALUATION_A,
                                             scenario_trade_fruit.VALUATION_B]
\end{Verbatim}

Similarly, scenarios are structured objects that can be used to generate more scenarios (i.e., new games). A \texttt{scenario} defines the initial context for a dialogue (e.g., an initial email).

\begin{Verbatim}[frame=leftline,fontsize=\footnotesize,numbers=left]
scenario_a = env_trade_fruit_with_tone_info.Scenario(
  scenario_trade_fruit.SCENARIO_A,
  "Bob",
  "Suzy",
  scenario_trade_fruit.ENDOWMENT_A,
  scenario_trade_fruit.VALUATION_A,
  "calm"
)

scenario_b = env_trade_fruit_with_tone_info.Scenario(
  scenario_trade_fruit.SCENARIO_B,
  "Jill",
  "George",
  scenario_trade_fruit.ENDOWMENT_B,
  scenario_trade_fruit.VALUATION_B,
  "calm"
)
 
examples_scenarios = [scenario_a, scenario_b]
\end{Verbatim}

Similar to observations and payoffs, an LLM termination prompt is a structured object that contains prompts for pre-processing the dialogue history using an LLM and then determining whether a given history is terminal by again prompting an LLM.

\begin{Verbatim}[frame=leftline,fontsize=\footnotesize,numbers=left]
llm_termination_prompt = scenario_trade_fruit.LLM_TERMINATION_PROMPT    
\end{Verbatim}

In addition to using an LLM, we can designate terminal histories ($\cZ$) by limiting the maximum number of replies per player (below). We can also specify the number of chance node outcomes (LLM seeds). Recall that the action space is combinatorial (recipient $\times$ tone) and so we define the number of actions accordingly. The \texttt{params} dictionary is passed to OpenSpiel to alert it to critical properties of the game that remain fixed (although \texttt{num\_llm\_seeds} and \texttt{num\_max\_replies} are specific to our \chatgames{}).

\begin{Verbatim}[frame=leftline,fontsize=\footnotesize,numbers=left]
params = {"num_distinct_actions": num_players * num_tones,
          "num_llm_seeds": 2,
          "num_players": num_players,
          "min_utility": min([float(p.min) for p in payoffs]),
          "max_utility": max([float(p.max) for p in payoffs]),
          "num_max_replies": 1}
config.params = params
\end{Verbatim}

Lastly, we incorporate these definitions into the configuration dictionary. Note that below, we are asking the LLM to generate $10$ names (each new game will randomly draw from this list) and $3$ of each type of private information (fruit endowment and valuation). The LLM list suffix is a simple modification to help the LLM generate new items given an initial list (e.g., of names).

\begin{Verbatim}[frame=leftline,fontsize=\footnotesize,numbers=left]
config.game = config_dict.ConfigDict()
config.game.observations = observations
config.game.header = header
config.game.payoffs = payoffs
config.game.given_prompt_actions = given_prompt_actions
config.game.num_names = 10
config.game.num_prompt_actions = (num_tones,)
config.game.num_private_info = (3, 3)
config.game.examples_names = examples_names
config.game.examples_private_info = examples_private_info
config.game.examples_scenarios = examples_scenarios
config.game.llm_list_suffix = "Output: "
config.game.llm_termination_prompt = llm_termination_prompt
\end{Verbatim}

This config dictionary defines the parameters of interaction between players in a dialogue in manner that binds precisely onto a game tree. We use configs like this one later in experiments to study three natural language settings: scheduling a meeting, trading fruit, and debate.

\end{document}